\documentclass[sigconf]{acmart}

\usepackage{stmaryrd}
\usepackage{amsmath}
\usepackage{amsfonts}

\SetSymbolFont{stmry}{bold}{U}{stmry}{m}{n}

\usepackage{ifthen}
\usepackage{algorithm}
\usepackage{algorithmic}

\newcommand{\mv}{\boldsymbol{m}}
\newcommand{\cov}{\boldsymbol{C}}

\newcommand{\x}{\boldsymbol{x}}
\newcommand{\y}{\boldsymbol{y}}
\newcommand{\z}{\boldsymbol{z}}

\newcommand{\E}{\mathbb{E}}

\newcommand{\R}{\mathbb{R}}
\newcommand{\X}{\mathcal{X}}

\newcommand{\N}{\mathcal{N}}
\newcommand{\argmax}{\mathop{\rm arg~max}\limits}
\newcommand{\argmin}{\mathop{\rm arg~min}\limits}

\newcommand{\T}{\mathrm{T}}

\newcommand{\I}{\mathbf{I}}

\newcommand{\p}{\boldsymbol{p}}

\usepackage{xcolor}
\usepackage{ifthen}
\newcommand{\markupdraft}[2]{
    \ifthenelse{\equal{#1}{display}}{#2}{}
    \ifthenelse{\equal{#1}{color}}{\color{#2}}{}
}

\newcommand{\newcolored}[3][]{{\markupdraft{color}{#2}#3}
    \ifthenelse{\equal{#1}{}}{}{\markupdraft{display}{{\color{yellow!70!black}[#1]}}}}
\newcommand{\del}[2][]{{\markupdraft{display}{{\color{orange}[removed: ``#2''[#1]]}}}} 
\newcommand{\new}[2][]{\newcolored[#1]{blue}{#2}}
\newcommand{\nnew}[2][]{\newcolored[#1]{red}{#2}}

\renewcommand{\del}[2]{}  
\renewcommand{\markupdraft}[2]{}  

\newcommand{\rev}[1]{#1}

\AtBeginDocument{%
  \providecommand\BibTeX{{%
    \normalfont B\kern-0.5em{\scshape i\kern-0.25em b}\kern-0.8em\TeX}}}

\copyrightyear{2024} 
\acmYear{2024} 
\setcopyright{acmlicensed}\acmConference[GECCO '24]{Genetic and Evolutionary Computation Conference}{July 14--18, 2024}{Melbourne, VIC, Australia}
\acmBooktitle{Genetic and Evolutionary Computation Conference (GECCO '24), July 14--18, 2024, Melbourne, VIC, Australia}
\acmDOI{10.1145/3638529.3654193}
\acmISBN{979-8-4007-0494-9/24/07}

\begin{document}

\title{CMA-ES for Safe Optimization}


\author{Kento Uchida}
\email{uchida-kento-fz@ynu.ac.jp}
\orcid{0000-0002-4179-6020}
\affiliation{%
  \institution{Yokohama National University}
  \city{Yokohama}
  \state{Kanagawa}
  \country{Japan}
  \postcode{240-8501}
}

\author{Ryoki Hamano}
\email{hamano_ryoki_xa@cyberagent.co.jp}
\orcid{0000-0002-4425-1683}
\affiliation{%
  \institution{CyberAgent, Inc. \and Yokohama National University}
  \city{Shibuya}
  \state{Tokyo}
  \country{Japan}
  \postcode{150-0042}
}

\author{Masahiro Nomura}
\email{nomura\_masahiro@cyberagent.co.jp}
\orcid{0000-0002-4945-5984}
\affiliation{%
  \institution{CyberAgent, Inc.}
  \city{Shibuya}
  \state{Tokyo}
  \country{Japan}
  \postcode{150-0042}
}

\author{Shota Saito}
\email{saito-shota-bt@ynu.jp}
\orcid{0000-0002-9863-6765}
\affiliation{%
  \institution{Yokohama National University \and SKILLUP NeXt Ltd.}
  \city{Yokohama}
  \state{Kanagawa}
  \country{Japan}
  \postcode{240-8501}
}

\author{Shinichi Shirakawa}
\email{shirakawa-shinichi-bg@ynu.ac.jp}
\orcid{0000-0002-4659-6108}
\affiliation{%
  \institution{Yokohama National University}
  \city{Yokohama}
  \state{Kanagawa}
  \country{Japan}
  \postcode{240-8501}
}

\renewcommand{\shortauthors}{K. Uchida et al.}

\begin{abstract}
In several real-world applications in medical and control engineering, there are unsafe solutions whose evaluations involve inherent risk. This optimization setting is known as safe optimization and formulated as a specialized type of constrained optimization problem with constraints for safety functions. Safe optimization requires performing efficient optimization without evaluating unsafe solutions. A few studies have proposed the optimization methods for safe optimization based on Bayesian optimization and the evolutionary algorithm. However, Bayesian optimization-based methods often struggle to achieve superior solutions, and the evolutionary algorithm-based method fails to effectively reduce unsafe evaluations. This study focuses on CMA-ES as an efficient evolutionary algorithm and proposes an optimization method termed safe CMA-ES. The safe CMA-ES is designed to achieve both safety and efficiency in safe optimization. The safe CMA-ES estimates the Lipschitz constants of safety functions transformed with the distribution parameters using the maximum norm of the gradient in Gaussian process regression. Subsequently, the safe CMA-ES projects the samples to the nearest point in the safe region constructed with the estimated Lipschitz constants. The numerical simulation using the benchmark functions shows that the safe CMA-ES successfully performs optimization, suppressing the unsafe evaluations, while the existing methods struggle to significantly reduce the unsafe evaluations.
\end{abstract}



\maketitle

\section{Introduction}
In the real-world applications within the medical and control engineering fields~\cite{safeopt:constraint,Louis:2022,Modugno:2016,safeopt}, unsafe solutions may be present, and their evaluations involve risk such as clinical deterioration and breakdown of control systems. 
For instance, in spinal cord therapy~\cite{safeopt}, the configuration of the electrical stimulation is optimized to improve spinal reflex and locomotor function, where unsafe configuration can aggravate spinal cord injuries. 
In real-world optimization of drone control system~\cite {safeopt:constraint}, we have to optimize the system parameters preventing drone collisions with the surrounding objects so as not to break down the drone.
These optimization problems, preventing the evaluation of unsafe solutions \rev{that should not be evaluated}, are termed as {\it safe optimization}. 
Safe optimization is formulated as a specialized type of constrained optimization problem aiming to reduce the evaluations of the solutions whose safety function values exceed the safety threshold.
Additionally, a set of safe solutions, referred to as {\it safe seeds} is provided to the optimizer.

\del{Several optimization methods have been developed to realize an efficient, safe optimization.}{}
\nnew{Several methods have been developed for safe optimization.}
SafeOpt~\cite{safeopt} is a representative in this category. SafeOpt relies on Bayesian optimization and constructs the safe region using the Lipschitz constant of the safety function not to evaluate the unsafe solutions.
Several extensions of SafeOpt have been proposed, such as {\it modified SafeOpt}~\cite{msafeopt}, which eliminates the need for the Lipschitz constant of the safety function, and {\it swarm-based SafeOpt}~\cite{safeoptswarm}, which introduces the particle swarm optimization~\cite{pso} to improve the search performance on high-dimensional problems.
In the context of evolutionary algorithms within safe optimization, a general approach called {\it violation avoidance}~\cite{Kaji:2009} has been proposed. This method involves regenerating a solution when the nearest solution to the generated one is deemed unsafe.
However, according to reference~\cite{Kim:2022}, it points out that the violation avoidance may not effectively suppress the evaluations of unsafe solutions compared with SafeOpt.
Considering the computational cost of updates and the \del{convergence speed of solutions}{}\rev{optimization performance with large budget}, evolutionary algorithms have advantages over Bayesian optimization~\rev{\cite{smac:bbob}}. Consequently, there is a demand for the development of efficient evolutionary algorithms tailored for safe optimization.

This study focuses on the covariance matrix adaptation evolution strategy (CMA-ES)~\cite{hansen:1996:ec} as an efficient evolutionary algorithm.
CMA-ES employs a multivariate Gaussian distribution as a sampling distribution for solutions and updates the distribution parameters to generate better solutions.
The CMA-ES possesses several invariance properties, such as the invariance to affine transformations of the search space, and realizes a powerful optimization performance on ill-conditioned and non-separable problems~\cite{hansen:2014:book}.

This study proposes an optimization method for safe optimization based on CMA-ES, termed safe CMA-ES, to realize both safety and efficiency in safe optimization.
In addition to the original update procedure of the CMA-ES, the safe CMA-ES estimates the Lipschitz constant of the safety function and constructs a safe region to repair the samples from \nnew{multivariate} Gaussian distribution.
The estimation process of the Lipschitz constant uses the Gaussian process regression (GPR)~\cite{rasmussen2006gaussian} trained with the evaluated solutions and computes the maximum norm of the gradient of the prediction mean.
The safe CMA-ES estimates the Lipschitz constant in the space transformed using the distribution parameter to inherit \nnew{the invariance to} affine transformations of the search space.
Then, the safe CMA-ES projects the sample generated from the \nnew{multivariate} Gaussian distribution to the nearest point in the safe region computed with the estimated Lipschitz constant.
Additionally, the safe CMA-ES corrects the initial distribution parameters using the safe seeds.

In numerical simulations, we evaluated the search performance of the safe CMA-ES on the benchmark functions for safe optimization.
While the existing method for safe optimization failed to suppress the evaluation of unsafe solutions, the safe CMA-ES successfully found the optimal solution with few or no unsafe evaluations.

\section{CMA-ES}

The CMA-ES is a probabilistic model-based evolutionary algorithm for continuous black-box optimization problems.
CMA-ES employs a multivariate Gaussian distribution parameterized by the mean vector $\mv \in \R^d$, the covariance matrix $\cov \in \R^{d \times d}$, and the step-size $\sigma \in \R_{>0}$.

We introduce the update procedure of the CMA-ES on the objective function $f:\R^d \to \R$.
At first, the CMA-ES generates $\lambda$ samples $\{ \x^{\langle \ell \rangle} \}_{\ell = 1}^\lambda$ from the multivariate Gaussian distribution \new{$\N(\mv^{(t)}, (\sigma^{(t)})^2 \cov^{(t)})$} as
\begin{align}
    \z^{\langle \ell \rangle} &\sim \N( \mathbf{0}, \I ) \label{eq:cma:z} \\
    \y^{\langle \ell \rangle} &= \sqrt{\cov^{(t)}} \z^{\langle \ell \rangle} \\
    \x^{\langle \ell \rangle} &= \mv^{(t)} + \sigma^{(t)} \y^{\langle \ell \rangle}
    \label{eq:cma:generation}
    \enspace.
\end{align}
Then, the CMA-ES evaluates the samples on the objective function and computes their rankings.
We denote the index of the $\ell$-th best sample as $\ell\!:\!\lambda$.

Next, the CMA-ES updates two evolution paths $\p_\sigma, \p_c \in \R^d$, which are initialized as $\p_\sigma^{(0)} = \p_c^{(0)} = \mathbf{0}$.
The update of evolution paths uses two weighted sums of $\mu$-best solutions $\Delta_{\z} = \sum_{\ell = 1}^{\mu} w_\ell \z^{\langle \ell:\lambda \rangle}$ and $\Delta_{\y} = \sum_{\ell = 1}^{\mu} w_\ell \y^{\langle \ell:\lambda \rangle}$ computed using the positive weights $w_1 \geq \cdots \geq w_\mu > 0$ and is performed as
\begin{align}
    \p_\sigma^{(t+1)} &= (1 - c_\sigma) \p_\sigma^{(t)} + \sqrt{ c_\sigma (2 - c_\sigma) \mu_\mathrm{eff} } \cdot \Delta_{\z} \\
    \p_c^{(t+1)} &= (1 - c_c) \p_c^{(t)} + h_\sigma^{(t+1)} \sqrt{ c_c (2 - c_c) \mu_\mathrm{eff} } \cdot \Delta_{\y}
    \enspace,
\end{align}
where $c_\sigma, c_c \in \R_{>0}$ are the accumulation rates of the evolution paths and $\mu_\mathrm{eff} = (\sum_{\ell=1}^\mu w_\ell^2)^{-1}$.
The Heaviside function $h_\sigma^{(t+1)} \in \{0, 1\}$ becomes one if and only if it satisfies 
\begin{align}
    \frac{\| \p_\sigma^{(t+1)} \| }{\sqrt{1 - (1 - c_\sigma)^{2 (t + 1)}}} < \left( 1.4 + \frac{2}{d+1} \right) \chi_d
    \enspace,
\end{align}
where $\chi_d = \sqrt{d} \left( 1 - \frac{1}{4d} + \frac{1}{21 d^2} \right)$ is approximated value of the expectation $\E[ \| \N(\mathbf{0}, \I) \|]$.

Finally, the CMA-ES updates the distribution parameters of the multivariate Gaussian distribution as
\begin{align}
    \mv^{(t+1)} &= \mv^{(t)} + c_m \sigma^{(t)} \Delta_{\y} 
    \label{eq:cma:update-mean} \\
    \sigma^{(t+1)} &= \sigma^{(t)} \exp \left( \frac{c_\sigma}{d_\sigma} \left( \frac{ \| \p_\sigma^{(t+1)} \| }{\chi_d} - 1 \right) \right)
    \label{eq:cma:update-stepsize}     
    \allowdisplaybreaks[3] \\
    \begin{split}
        \cov^{(t+1)} &= (1 + (1 - h_\sigma^{(t+1)}) c_1 c_c (2 - c_c)) \cov^{(t)} \\
        & + c_1 \left( \p_c^{(t+1)} \left( \p_c^{(t+1)} \right)^\T - \cov^{(t)} \right) \\
        & + c_\mu \sum_{\ell = 1}^{\mu} w_\ell \left( \y^{\langle \ell:\lambda \rangle} \left( \y^{\langle \ell:\lambda \rangle} \right)^\T - \cov^{(t)} \right)
    \end{split}
    \label{eq:cma:update-cov} 
\end{align}
where $c_m, c_1, c_\mu \in \R_{>0}$ are the learning rates and $d_\sigma \in \R_{>0}$ is the damping factor.
The CMA-ES contains well-tuned default values for hyperparameters. 
Refer to the details in the references~\cite{hansen:2017:arxiv,hansen:2014:book}.

\del{
\subsection{Gaussian Process Regression}
The Gaussian process regression (GPR) is a nonparametric regression method which expresses the prediction by the posterior distribution of the function following a Gaussian process.
The Gaussian process is a set of random variables where the law of $n$ evaluation values $(g(\x_1), \cdots, g(\x_n))^\T$ is given by a multivariate Gaussian distribution.
The Gaussian process is determined by the kernel function $k: \R^d \times \R^d \to \R$ that gives the multivariate Gaussian distribution.

We consider a function $g: \R^d \to \R$ given by the Gaussian process.
When we have $n$ observations $\mathcal{D} = \{ (\boldsymbol{x}_i, g(\x_i)) \}_{i=1}^n$ on $g$, the posterior distribution of evaluation value $g(\x_{\mathrm{new}})$ at the target input $\x_{\mathrm{new}}$ is given by the Gaussian distribution with the mean $\mu(\boldsymbol{x}_{\mathrm{new}} \mid \mathcal{D})$ and variance $\sigma^2(\boldsymbol{x}_{\mathrm{new}} \mid \mathcal{D})$ as
\begin{align}
    \mu(\boldsymbol{x}_{\mathrm{new}} \mid \mathcal{D}) &= \boldsymbol{k}_{\ast}^\T \boldsymbol{K}^{-1} \boldsymbol{t} \label{eq:gp:mean} \\
    \sigma^2(\boldsymbol{x}_{\mathrm{new}} \mid \mathcal{D}) &= k_{\ast \ast} - \boldsymbol{k}_{\ast}^\T \boldsymbol{K}^{-1} \boldsymbol{k}_{\ast}
    \enspace,
\end{align}
where $\boldsymbol{k}_{\ast} \in \mathbb{R}^{n}$, $k_{\ast \ast} \in \mathbb{R}$ and $\boldsymbol{t} \in \mathbb{R}^{n}$ is given by
\begin{align*}
    \boldsymbol{k}_{\ast} &= \left( 
        k(\boldsymbol{x}_1, \boldsymbol{x}_{\mathrm{new}}), \cdots, k(\boldsymbol{x}_n, \boldsymbol{x}_{\mathrm{new}}) \right) \\
    k_{\ast \ast} &= k(\boldsymbol{x}_{\mathrm{new}}, \boldsymbol{x}_{\mathrm{new}}) \\
    \boldsymbol{t} &= (g(\x_1), \cdots, g(\x_n))^\T
    \enspace.
\end{align*}
The matrix $\boldsymbol{K}$ is Gram matrix whose element is given by $\boldsymbol{K}_{i,j} = k(\boldsymbol{x}_i, \boldsymbol{x}_j)$.
The GPR has been used as a surrogate model for the CMA-ES~\cite{gpcmaes, dtscmaes}, which implies that the CMA-ES works along well with the GPR.
}{}

\begin{algorithm*}[!t] 
\centering
\caption{Safe CMA-ES}
\begin{algorithmic}[1] \label{alg:safecmaes}
\REQUIRE The objective function $f$ to be minimized
\REQUIRE Initial distribution parameters $\mv^{(0)}, \cov^{(0)}, \sigma^{(0)}$ 
\REQUIRE \nnew{Hyperparameters for estimation of Lipschitz constant: $T_{\mathrm{data}} = 5$, $\zeta_{\mathrm{init}} = 10$, $\alpha = 10$}
\REQUIRE \nnew{Hyperparameters for initialization: $L_{\min} = 100$, $\gamma = 0.9$}
\REQUIRE Safe seeds $\X_\mathrm{seed} = \{ \x_\mathrm{seed} \}$, safety thresholds $\{ h_j \}_{j=1}^p$ iterator $t=0$
\STATE Compute initial estimation of the Lipschitz constants $L_1^{(0)}, \cdots, L_p^{(0)}$ of the transformed safety functions using (\ref{eq:init:lip}).
\STATE Modify the initial mean vector $\mv^{(0)}$ and the initial step-size $\sigma^{(0)}$ using (\ref{eq:init:mean}) and (\ref{eq:init:stepsize}), respectively.
\WHILE{termination condition is not met}
\FOR{$\ell = 1, \cdots, \lambda$}
\STATE Generate $\z^{\langle \ell \rangle} \sim \N(\mathbf{0}, \I)$ and project $\z^{\langle \ell \rangle}$ to the nearest point $\tilde{\z}^{\langle \ell \rangle}$ in safe region by (\ref{eq:modify:z}).
\STATE Compute $\x^{\langle \ell \rangle}$ using (\ref{eq:cma:generation}) with modified sample $\tilde{\z}^{\langle \ell \rangle}$.
\STATE Evaluate $\x^{\langle \ell \rangle}$ on the objective function $f$ and the safety functions $s_j$ for $j = 1, \cdots, p$.
\ENDFOR
\STATE Update $\mv^{(t+1)}, \cov^{(t+1)}, \sigma^{(t+1)}$ \rev{with modified samples $\{ \x^{\langle \ell \rangle} \}_{\ell = 1}^\lambda$} using the update rules (\ref{eq:cma:update-mean}), (\ref{eq:cma:update-stepsize}), and (\ref{eq:cma:update-cov}), respectively.
\STATE Estimate the Lipschitz constants $L_1^{(t+1)}, \cdots, L_p^{(t+1)}$ of the transformed safety functions using (\ref{eq:lip-est}) \rev{with $N_\mathrm{data}$ solutions evaluated in the last $T_\mathrm{data}$ iterations}.
\STATE $t \leftarrow t + 1$
\ENDWHILE
\end{algorithmic} 
\end{algorithm*}
%
%
%

\section{Formulation of Safe Optimization}
We follow the formulation of the safe optimization outlined in the reference~\cite{survey}.
We consider a constrained minimization problem of the objective function as
\begin{align}
    \begin{split}
        \min_{\x \in \R^d} \, f(\x) \quad
        \text{s.t.} \quad s_j(\x) \leq h_j  \quad \text{for all} \enspace j=1, \cdots, p 
    \end{split}
\end{align}
where $s_j: \R^d \to \R$ and $h_j \in \R$ are the safety function and the safety threshold, respectively.
We consider a solution as a {\it safe solution} when it satisfies all safety constraints and regard a solution as an {\it unsafe solution} when it violates at least one safety constraint.
In safe optimization, the optimizer can access the safety thresholds and is required to optimize the objective function, suppressing the evaluations of the unsafe solutions.
In addition, the optimizer receives $N_\mathrm{seed}$ safe solutions as the safe seeds.
The upper limit of unsafe evaluations depends on the target application.
The evaluation of unsafe solutions is usually prohibited in medical applications, while a few unsafe evaluations may be acceptable for control system optimization.
For example, in control system optimization for robots, one can prepare multiple robots of the same type in preparation for breakage, and the number of robots serves as the upper limit for unsafe evaluations in this case.

We assume that the evaluations of the objective function and the safety constraints are jointly performed.
Additionally, we assume that the safety functions are noiseless black-box functions.
For simplicity in this research, we assume that the evaluation values of the unsafe solutions on the objective function are accessible.

\section{Proposed Method: Safe CMA-ES}
This study proposes the safe CMA-ES, an extension of the CMA-ES that achieves efficient optimization performance in safe optimization.
\del{
The safe CMA-ES constructs the safe region based on the estimated Lipschitz constant of the safety functions transformed using distribution parameters.
Then, the safe CMA-ES projects the generated samples to the nearest point in the safe region so as not to evaluate the unsafe solutions.
}{}
\nnew{
The safe CMA-ES constructs the safe region based on the estimated Lipschitz constants of the safety functions and projects the generated samples to the nearest point in the safe region to avoid evaluating unsafe solutions (Section~\ref{sec:proposed:projection}).
The Lipschitz constants are estimated in the transformed search space by the Gaussian process regression (Section~\ref{sec:proposed:lipschitz}).
}
To further enhance safety, the safe CMA-ES initializes the distribution, ensuring that it fits within the safe region (Section~\ref{sec:proposed:init}).
Algorithm~\ref{alg:safecmaes} shows the pseudo-code of the safe CMA-ES.
\begin{figure*}[!t]
  \centering
  \includegraphics[width=0.8\linewidth]{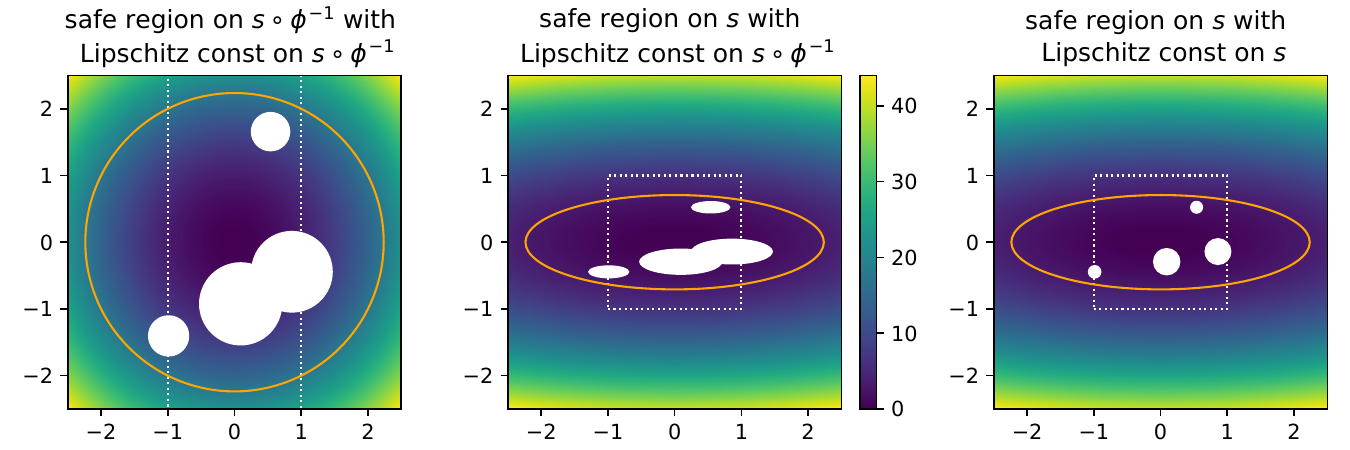}
  \caption{The safe region on the safety function $s(\x) = x_1^2 + 10 x_2^2$ with the safety threshold $h=5$ with four safe seeds. The distribution parameters are given by $\mv = \mathbf{0}$, $\sigma=1$, and $\cov= \mathrm{diag}((1, 0.1)^\T)$. We generated safe seeds uniformly at random in the range $[-1,1]^2$ plotted as a white dotted box. The white circles and the orange line show the safe region and the border of the safety constraint. The left figure shows the safe region on the composition function $s_j \circ \phi^{-1}$, and the center figure shows the safe region on the safety function $s_j$. The right figure shows the safe region on the safety function $s_j$ computed with the Lipschitz constant of the safety function $s_j$ instead of the composition function $s_j \circ \phi^{-1}$.}
  \label{fig:safe_region}
\end{figure*}
%
%
%

\subsection{Projection of Samples to Safe Region}
\label{sec:proposed:projection}
The safe CMA-ES introduces the transformation $\phi$ using the current distribution parameters $\boldsymbol{\theta}^{(t)} = \{ \mv^{(t)}, \cov^{(t)}, \sigma^{(t)} \}$ to inherit the invariance to affine transformations of the search space as
\begin{align}
     \phi(\x) := \phi(\x; \boldsymbol{\theta}^{(t)}) = \left( \sqrt{\cov^{(t)}} \right)^{-1} \frac{\x - \mv^{(t)}}{\sigma^{(t)}} \enspace.
\end{align}

The safe CMA-ES constructs the safe region using the following fact: given the Lipschitz constant $L_j$ of the composition function $s_j \circ \phi^{-1}$, any safe solution $\x \in \R^d$ and any solutions $\x' \in \R^d$ satisfy
\begin{align}
    \| \phi(\x) - \phi(\x') \| &\leq \frac{ h_j - s_j(\x) }{ L_j } 
    \enspace \Rightarrow \enspace
    s_j(\x') \leq h_j
    \enspace.
\end{align}
The safe CMA-ES uses $N_\mathrm{data} = \min\{N_\mathrm{seed} + \lambda t, \lambda T_\mathrm{data} \}$ solutions evaluated in the last $T_\mathrm{data}$ iterations to construct the safe region.
The safe CMA-ES computes the safe region using safe solutions $\mathcal{A}_\mathrm{safe}$ in $N_\mathrm{data}$ evaluated solutions and the estimated Lipschitz constant $L_j^{(t)}$, explained in the next section, as
\begin{align*}
    \mathcal{X}_\mathrm{safe} = \bigcup_{\x \in \mathcal{A}_\mathrm{safe}} \left\{ \x' \in \X \mid \| \phi(\x) - \phi(\x') \| \leq \delta(\x)\right\} 
    \enspace.
\end{align*}
The function $\delta$ returns the radius of the safe region with a given safe solution on the composition function $s_j \circ \phi^{-1}$, as
\begin{align}
    \delta(\x) &= \min_{1 \leq j \leq p} \frac{h_j - s_j(\x)}{ L_j^{(t)} } 
    \enspace.
    \label{eq:slack} 
\end{align}
Figure~\ref{fig:safe_region} illustrates an example of the safe region.
Figure~\ref{fig:safe_region} also shows the safe region computed with the Lipschitz constant of the safety function $s_j$ instead of the composition function $s_j \circ \phi^{-1}$ for reference.
It can be observed that the Lipschitz constant of our composition function expands the safe region \rev{when the distribution parameters of the CMA-ES is learned appropriately}.

To prevent the evaluation of unsafe solutions, the safe CMA-ES projects the generated solution $\x^{\langle \ell \rangle}$ in~\eqref{eq:cma:generation} to the nearest point in the safe region with respect to the Mahalanobis distance as
\begin{align}
    \tilde{\x}^{\langle \ell \rangle} \in \argmin_{\x' \in \mathcal{X}_{\mathrm{safe}}} \| \phi(\x') - \phi(\x^{\langle \ell \rangle}) \| 
    \label{eq:modify:x}
    \enspace.
\end{align}
The projection is performed by modifying the sample $\z^{\langle \ell \rangle}$ generated in~\eqref{eq:cma:z} as
\begin{align}
    \tilde{\z}^{\langle \ell \rangle} =  \xi^{\langle \ell \rangle} \z^{\langle \ell \rangle} + (1 - \xi^{\langle \ell \rangle}) \phi(\x^{\langle \ell \rangle}_\mathrm{near})
    \enspace,
    \label{eq:modify:z}
\end{align}
where $\x^{\langle \ell \rangle}_\mathrm{near} \in \mathcal{A}_{\mathrm{safe}}$ and $\xi^{\langle \ell \rangle} \in [0,1]$ are given as
\begin{align}
    \x^{\langle \ell \rangle}_\mathrm{near} &= \argmax_{\x \in \mathcal{A}_\mathrm{safe}} \left\{ \delta(\x) - \| {\z}^{\langle \ell \rangle} - \phi(\x) \| \right\} \\
    \xi^{\langle \ell \rangle} &= \min\left\{1, \frac{ \delta(\x^{\langle \ell \rangle}_\mathrm{near}) }{ \| {\z}^{\langle \ell \rangle} - \phi(\x^{\langle \ell \rangle}_\mathrm{near}) \| } \right\} \enspace.
\end{align}
\new{
The safe solution $\x^{\langle \ell \rangle}_\mathrm{near}$ has the safe region closest to ${\z}^{\langle \ell \rangle}$ in $\mathcal{A}_\mathrm{safe}$, and $\xi^{\langle \ell \rangle}$ determines $\tilde{\z}^{\langle \ell \rangle}$ between ${\z}^{\langle \ell \rangle}$ and $\phi(\x^{\langle \ell \rangle}_\mathrm{near})$, which becomes $\xi^{\langle \ell \rangle} = 1$ when the solution $\x^{\langle \ell \rangle}$ is originally generated in the safe region.
}

\begin{figure*}[t]
    \centering
    \includegraphics[width=0.85\linewidth]{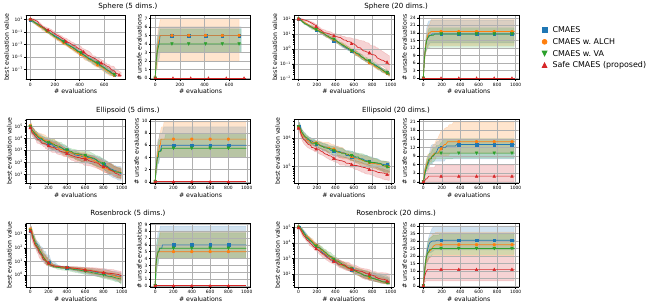}
    \caption{Transitions of the best evaluation value and the number of evaluations of unsafe solutions with safety function $s(\x) = f(\x)$. We plot the medians and interquartile ranges over 50 trials.}
    \label{fig:exp1}
\end{figure*}
%
%
%

\subsection{Estimation of Lipschitz Constants}
\label{sec:proposed:lipschitz}
Based on the updated distribution parameters, the safe CMA-ES estimates the Lipschitz constant of each safety function using the \nnew{Gaussian process
regression (GPR)} trained with evaluated solutions.
The safe CMA-ES also uses $N_\mathrm{data}$ solutions evaluated in the last $T_\mathrm{data}$ iterations (including the solutions evaluated in the current iteration) as the training data for the GPR.
The safe CMA-ES normalizes the $i$-th solution in the training data $\x_i$ using the updated distribution parameters as $\z_i = \phi(\x_i; \boldsymbol{\theta}^{(t+1)})$.
Additionally, the target variable corresponding to the $i$-th training data is normalized as
\begin{align}
    \omega_{i,j} = \frac{s_j(\x_i) - \mu_j }{\sigma_j} \enspace,
\end{align}
where $\mu_j$ and $\sigma_j$ are the average and standard deviation of the evaluation values on the $j$-th safety function over the $N_\mathrm{data}$ solutions, respectively.
The safe CMA-ES computes the estimated Lipschitz constant of the composition function $s_j \circ \phi^{-1}$ using the posterior distribution of GPR with the training data $\mathcal{D} = \{(\z_i, \omega_{i, j})\}_{i=1}^{N_\mathrm{data}}$ as
\begin{align}
    \hat{L}_j^{(t+1)} = \sigma_j \cdot \max_{\z \in \mathcal{Z}} \left\| \nabla_{\z} \mu(\z \mid \mathcal{D}) \right\|
    \enspace,
    \label{eq:lip-est-init}
\end{align}
where $\mu(\z \mid \mathcal{D})$ is the mean of the posterior distribution.
We set the search space for maximization in \eqref{eq:lip-est-init} as $\mathcal{Z} = [-3, 3]^d$.
Based on the approach in~\cite{Gonzalez:2016}, the safe CMA-ES applies a two-step optimization process to solve the maximization.
In the first step, the safe CMA-ES generates $5 \lambda$ samples $\hat{\mathcal{Z}} = \{ \hat{\z}_i \}_{i=1}^{5 \lambda}$ from the standard multivariate \nnew{normal} distribution $\mathcal{N}(\mathbf{0}, \I)$ and computes the norm $\| \nabla_{\z} \mu(\z) \|$ of the gradient of the mean of the posterior distribution with respect to each sample $\hat{\z}_i$.
In the second step, the safe CMA-ES employs L-BFGS~\cite{lbfgs} with box-constraint handling and runs L-BFGS from the maximizer in $\hat{\mathcal{Z}}$ for 200 iterations.
Following the settings of the surrogate-assisted CMA-ES with GPR~\cite{gpcmaes, sgpcmaes}, we use the RBF kernel as
\begin{align}
    k(\z, \z') = \exp \left( \frac{\| \z - \z' \|^2}{2 H^2} \right) \enspace. \label{eq:rbf}
\end{align}
We set the length scale and observation noise as $H = 8d$ and $0$, respectively.

In addition, the safe CMA-ES has two correction mechanisms \del{of}{}\nnew{that increase} the estimated Lipschitz constant in case the estimation using the GPR is unreliable.
\new{
Introducing coefficients $\tau^{(t+1)}$ and $\rho_j^{(t+1)}$ for the correction, the estimated Lipschitz constant is computed as
\begin{align}
    L_j^{(t+1)} = \tilde{L}_j^{(t+1)} \cdot \tau^{(t+1)} \cdot \rho_j^{(t+1)} 
    \enspace.
    \label{eq:lip-est}
\end{align}
The updates of those coefficients are as follows.
}
\del{As the first correction mechanism, because the prediction with small training data set is unreliable, the safe CMA-ES introduces a coefficient $\tau^{(t+1)}$ for the estimated Lipschitz constant as}{}
\new{
The safe CMA-ES increases the coefficient $\tau^{(t+1)}$ when the number of training data $N_\mathrm{data}$ is small because the prediction with small training dataset is unreliable. 
\del{Remaining}{}\nnew{Recalling} the maximum number of training data is $\lambda T_\mathrm{data}$, the coefficient $\tau^{(t+1)}$ is updated as
}
\begin{align}
    \tau^{(t+1)} = \begin{cases}
        (\zeta_\mathrm{init})^{\frac{1}{N_\mathrm{data}}} & \text{if} \enspace N_\mathrm{data} < \lambda T_\mathrm{data} \\
        1 & \text{otherwise}
    \end{cases}
    \enspace,
\end{align}
where $\zeta_\mathrm{init} > 1$ determines the effect of the \del{first }{}coefficient \nnew{$\tau^{(t+1)}$}.
\del{As the second correction mechanism, the safe CMA-ES introduces another coefficient $\rho_j^{(t+1)}$ for each safety constraint which is increased when the solutions violate it.
The update of the coefficient depends on the ratio $\nu_j$ of unsafe solutions in $\lambda$ solutions generated in the current iteration as}{}
\new{
The coefficient $\rho_j^{(t+1)}$ is set for each safety constraint and increased when the solutions violate it.
Since the safety constraints violated by many solutions require drastic correction, the safe CMA-ES determines the update strength of the coefficient based on the ratio $\nu_j$ of $\lambda$ solutions generated in the current iteration as
}
\begin{align}
    \nu_j =  \frac{1}{\lambda} \sum_{i=1}^{\lambda} \mathbb{I}\{ s_j(\x_i) > h_j \}
    \enspace.
\end{align}
Then, the safe CMA-ES updates the coefficient as
\begin{align}
    \rho_j^{(t+1)} =  \begin{cases}
        \rho_j^{(t)} \cdot \alpha^{\nu_j} & \text{if} \quad \nu_j > 0 \\
        \max\{ 1, \rho_j^{(t)} / \alpha^{1 / d} \} & \text{otherwise} 
    \end{cases}
    \enspace
\end{align}
where $\alpha > 1$ determines the effect of the \del{second }{}coefficient \nnew{$\rho_j^{(t+1)}$}.
We set the initial value of the coefficient as $\rho_j^{(0)} = 1$.

\del{Finally, the estimated Lipschitz constant is computed as
\begin{align}
    L_j^{(t+1)} = \tilde{L}_j^{(t+1)} \cdot \tau^{(t+1)} \cdot \rho_j^{(t+1)} 
    \enspace.
    \label{eq:lip-est}
\end{align}
}{}

\del{
\begin{table}[t]
\centering
\caption{Recommended hyperparameters of the safe CMA-ES.}
\begin{tabular}{c}
\hline
    hyperparameters for estimation of Lipschitz constant: \\
    $T_{\mathrm{data}} = 5$ , \hspace{10pt} $\zeta_{\mathrm{init}} = 10$ , \hspace{10pt} $\alpha = 10$ \\
\hline
\hline
    hyperparameters for initialization: \\
    $L_{\min} = 100$ , \hspace{10pt} $\gamma = 0.9$ \\
\hline
\end{tabular}
\label{table:hyperparameter}
\end{table}
}{}

\subsection{Initialization of Distribution Parameters}
\label{sec:proposed:init}
The Lipschitz constants of the safety functions are estimated before generating solutions in the first iteration.
In the first estimation, the safe CMA-ES \del{set}{}\nnew{requires} a \nnew{predefined} lower bound $L_{\min}$ for the estimated Lipschitz constant and estimates the Lipschitz constant as
\begin{align}
    L_j^{(0)} = \max\left\{ L_\mathrm{min}, \hat{L}^{(0)}_j \cdot \tau^{(0)} \right\} 
    \enspace.
    \label{eq:init:lip}
\end{align}
We note $\hat{L}_j^{(0)}$ is computed with the initial distribution parameters $\mv^{(0)}, \sigma^{(0)}, \cov^{(0)}$ and the GPR with the safe seeds, and $\tau^{(0)} = (\zeta_\mathrm{init})^{1 / {N_\mathrm{seed}}}$ is computed with the number $N_\mathrm{seed}$ of the safe seeds.
If the number of the safe seeds is one, the safe CMA-ES set $L_j^{(0)} = L_\mathrm{min}$ because the GPR does not work well.

The safe CMA-ES also modifies the initial mean vector and step-size using the safe seeds.
The initial mean vector is set to the safe seed with the best evaluation value on the objective function as
\begin{align}
    \tilde{\mv}^{(0)} = \argmin_{\x \in \X_{\mathrm{seed}}} f(\x) \enspace. \label{eq:init:mean}
\end{align}
The step-size is modified to maintain the ratio of solutions not changed after the projection in~\eqref{eq:modify:z} above $\gamma \in (0, 1)$ as
\del{
\begin{align}
    \tilde{\sigma}^{(0)} = \sigma^{(0)} \cdot \min \left\{ \frac{ \delta( \tilde{\mv}^{(0)} ) }{ \chi_{\mathrm{ppf}}( (1 + \gamma^{1 / d})/2 ) } , \, 1 \right\} 
    \enspace,
    \label{eq:init:stepsize}
\end{align}
where $\chi_{\mathrm{ppf}}$ is the inverse function of the cumulative density function of the standard Gaussian distribution. 
}{}
\new{
\begin{align}
    \tilde{\sigma}^{(0)} = \sigma^{(0)} \cdot \min \left\{ \frac{ \delta( \tilde{\mv}^{(0)} ) }{ \sqrt{ \chi^2_{\mathrm{ppf}}( \gamma ) } } , \, 1 \right\} 
    \enspace,
    \label{eq:init:stepsize}
\end{align}
where $\chi^2_{\mathrm{ppf}}$ is the percent point function of the $\chi^2$-distribution with the degree of freedom $d$, \nnew{which gives the squared radius of trust region on the standard $d$-dimensional Gaussian distribution with the probability $\gamma$}. 
}
We note the function $\delta$ is given by~\eqref{eq:slack}.
We set the target ratio to $\gamma = 0.9$.


\del{
\subsection{Recommended Hyperparameters}
Table~\ref{table:hyperparameter} shows the recommended setting of the hyperparameters for the safe CMA-ES.
We investigate the sensitivity of hyperparameters in the supplementary material.
}{}

\begin{figure*}[t]
    \centering
    \includegraphics[width=0.85\linewidth]{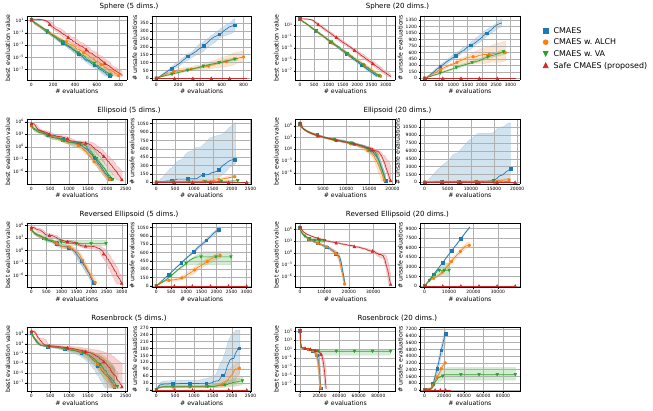}
    \caption{Transitions of the best evaluation value and the number of evaluations of unsafe solutions with safety function $s(\x) = x_1$. We plot the medians and interquartile ranges over 50 trials.}
    \label{fig:exp2}
\end{figure*}
%
%
%

\section{Experiment}
We investigate the following aspects through numerical simulation.\footnote{The code will be made available at \textcolor{blue}{\url{https://github.com/CyberAgentAILab/cmaes}}~\cite{nomura2024cmaes}.}
\begin{itemize}
\setlength{\leftskip}{-0.35cm}
    \item The performance evaluation of the safe CMA-ES in the early phase of the optimization (Section~\ref{sec:exp_early}).
    \item The performance evaluation of the safe CMA-ES throughout the optimization (Section~\ref{sec:exp_throughout}).
    \item The performance comparison of the safe CMA-ES with the optimization methods for safe optimization (Section~\ref{sec:exp_comparison}).
\end{itemize}
\rev{We also provide the results of sensitivity experiments of hyperparameters of the safe CMA-ES in the supplemental material.}

\subsection{Comparative Methods}
\paragraph{SafeOpt}
SafeOpt~\cite{safeopt} is an optimization method for safe optimization based on Bayesian optimization.
SafeOpt assumes accessibility to the Lipschitz constant $L$ of the safety function and constructs the safe region based on the upper confidence bound $u_t(\x)$ of the evaluation value \nnew{on the safety function} as
\footnote{Differently from the original paper~\cite{safeopt}, which assumes the safety constraint $s(\x) \geq h$, we explain the update of SafeOpt with the safety constraint $s(\x) \leq h$.}
\begin{align}
    S_t = \bigcup_{\x \in S_{t-1}} \left\{ \x' \in \X \mid \| \x - \x' \| \leq \frac{h - u_{t}(\x)}{L} \right\}
\end{align}
The initial safe region is given by the safe seeds.
Subsequently, SafeOpt computes two regions within the safe region: the region $M_t \subseteq S_t$, where the optimal solution seems to be included, and the region $G_t \subseteq S_t$, comprising solutions that could potentially extend the safe region.
The SafeOpt evaluates the solution with the largest confidence interval in the union $M_t \cup G_t$.
As a variant of SafeOpt, reference~\cite{safeopt} also proposes SafeOpt-UCB, which evaluates the solution with the smallest lower bound of the predictive confidence interval within the safe region $S_t$.

Several methods have been developed as extensions of SafeOpt.
The reference~\cite{msafeopt} modified the update of SafeOpt not to require the Lipschitz constant of the safety function and proposed {\it modified SafeOpt.}
Additionally, to reduce the computational cost in high-dimensional problems, swarm-based SafeOpt~\cite{safeoptswarm} uses the particle swarm optimization~\cite{pso} to search for the solution on the GPR.

\paragraph{Violation Avoidance}
The violation avoidance~\cite{Kaji:2009} is a general handling for evolutionary algorithms in the safe optimization.
The violation avoidance modifies the generation process of the evolutionary algorithm to regenerate a solution when the nearest solution to the generated one is unsafe.
The distance between the generated solution $\x_\mathrm{new}$ and the solution $\x_\mathrm{old}$ already evaluated is given by
\begin{align}
    d(\x_\mathrm{new}, \x_\mathrm{old}) = \frac{1}{w(\x_\mathrm{old})} \cdot \| \x_\mathrm{new} - \x_\mathrm{old} \|
    \enspace.
\end{align}
The weight is determined based on whether $\x_\mathrm{old}$ is safe or unsafe as
\begin{align}
    w(\x_\mathrm{old}) = \begin{cases}
    w_\mathrm{safe} & \text{if $\x_\mathrm{old}$ is safe} \\ 
    w_\mathrm{unsafe} & \text{if $\x_\mathrm{old}$ is unsafe} 
    \end{cases}
\end{align}
where $w_\mathrm{safe}, w_\mathrm{unsafe} \in \R_{>0}$ are the predefined weights.
The evolutionary algorithm tends to evaluate a solution close to a safe solution when $w_\mathrm{safe}$ is larger than $w_\mathrm{unsafe}$ and close to an unsafe solutions otherwise.

In numerical simulation, we included the CMA-ES with violation avoidance as one of the comparative methods.
We set the weights as $w_\mathrm{safe} = w_\mathrm{unsafe} = 1$.
We sampled $10 \lambda$ points from the multivariate Gaussian distribution and randomly selected $\lambda$ solutions from the samples whose closest evaluated solutions are safe.
We terminated the optimization when we could not obtain $\lambda$ solutions from $10 \lambda$ samples.

\subsection{Experimental Setting}
We used the following benchmark functions with a unique optimal solution $\x^\ast = \mathbf{0}$.
\begin{itemize}
\setlength{\leftskip}{-0.35cm}
    \item Sphere: $f(\x) = \sum_{i=1}^d x_i^2$
    \item Ellipsoid: $f(\x) = \sum_{i=1}^d \left( 1000^{\frac{i-1}{d-1}} x_i \right)^2$
    \item Reversed Ellipsoid: $f(\x) = \sum_{i=1}^d \left( 1000^{\frac{d-i}{d-1}} x_i \right)^2$
    \item Rosenbrock: $f(\x) = \sum_{i=1}^{d-1} \left( 100 ((x_{i+1} + 1) - (x_i + 1)^2)^2 + x_i^2 \right)$
\end{itemize}
The sphere, ellipsoid, and rosenbrock functions are well-known benchmarks for continuous black-box optimization.
Additionally, we designed the reversed ellipsoid function by reversing the order of the coefficients in the ellipsoid function.
This reversed ellipsoid function was employed to investigate the impact of safety function settings on the performance of the safe CMA-ES.

In the first and second experiments, we compared the safe CMA-ES with three comparative methods: the naive CMA-ES, CMA-ES with constraint handling, and CMA-ES with violation avoidance.
We used augmented Lagrangian constraint handling~\cite{alch:cmaes}\footnote{\new{We implemented the constraint handling with \texttt{pycma}~\cite{hansen:pycma:2019}. We do not use the negative weights for fair comparison.}} as the constraint handling for the CMA-ES\del{, implemented with \texttt{pycma}~\cite{hansen:pycma:2019}}{}.
We terminated the optimization when the best evaluation value reached $10^{-8}$ or when the minimum eigenvalue of $(\sigma^{(t)})^2 \cov^{(t)}$ became smaller than $10^{-30}$.

In the third experiment, we compared the safe CMA-ES with the existing optimization methods designed for safe optimization: SafeOpt, modified SafeOpt, and swarm-based SafeOpt.\footnote{We used the implementation of SafeOpt, modified SafeOpt, and swarm-based SafeOpt provided by authors in \url{https://github.com/befelix/SafeOpt}.}
We used the RBF kernel in~\eqref{eq:rbf} as the kernel of them for a fair comparison with the safe CMA-ES.%
\footnote{
We did not use the automatic relevance determination (ARD) as it did not lead performance improvements in preliminary experiments.
}
We optimized the hyperparameters of the kernel by the maximum likelihood method.
As SafeOpt requires the Lipschitz constant of the safety function, we provided the maximum of the norm of the gradient over the grid points that divide each dimension on the search space into $11$ equally, i.e., $11^d$ points on the grid in total.
For the violation avoidance, consistent with the setting in the original paper, we fixed $w_\mathrm{unsafe}=1$ and varied the weight for safe solution \nnew{in addition to the default setting} as $w_\mathrm{safe}=0.5, 1, 2$.

For all methods, we set the search space as $\X = [-5, 5]^d$.
We obtained the safe seeds from the samples that are generated uniformly at random within the search space and satisfy all safety constraints.
We set the number of the safe seeds as $N_\mathrm{seed} = 10$.
For the methods employing CMA-ES, the initial mean vector $\mv^{(0)}$ was set to the safe seed with the best evaluation value on the objective function.
The initial step-size and the covariance matrix are given by $\sigma^{(0)} = 2$ and $\cov^{(0)} = \I$, respectively.
It is important to note that the safe CMA-ES corrects the initial step-size using~\eqref{eq:init:stepsize}.
We set the number of dimensions as $d=5, 20$.
\new{The population size was set as $\lambda = 4 + \lfloor 3 \ln d \rfloor$.}
We conducted 50 independent trials for each setting.

\subsection{Result of Performance Evaluation in Early Phase of Optimization}
\label{sec:exp_early}
We set the safety function and threshold to investigate the performance in the early phase of the optimization process as
\begin{align}
    s(\x) = f(\x) \qquad \text{and} \qquad h = q(f, \X, 0.5)
    \enspace,
    \label{eq:exp1:constraint}
\end{align}
where $q(f, \X, \gamma)$ represents $\gamma$-quantile of the uniform distribution on the objective function $f$ over the search space $\X$. 
In this case, the solution with a poor evaluation value on the objective function is unsafe.
The safety threshold $q(f, \X, \gamma)$ was estimated using 10,000 samples generated uniformly at random across the search space.
We set the total number of evaluations to 1,000.

Figure~\ref{fig:exp1} shows the transitions of the best evaluation value and the number of unsafe evaluations.
Focusing on the result of 5-dimensional problems, the safe CMA-ES successfully optimized all benchmark functions, suppressing the unsafe evaluations. 
It is noteworthy that the safe CMA-ES avoided evaluating unsafe solutions in over 75\% of the trials.
Meanwhile, the violation avoidance did not significantly reduce the unsafe evaluations compared to the naive CMA-ES.

For 20-dimensional problems, the safe CMA-ES initially evaluated the unsafe solutions in the early iterations on the ellipsoid and rosenbrock functions.
We consider that this is because the estimation of 20-dimensional GPR trained with limited training data was unreliable.
The safe CMA-ES successfully evaluated only safe solutions after initial iterations.
Meanwhile, we observe a gradual decrease in the best evaluation value of the safe CMA-ES on the sphere function compared to other methods.
We consider that the overestimation of the Lipschitz constant of the safety function led to this slight deterioration. 
However, the safe CMA-ES accelerated the decrease of the best evaluation value on the ellipsoid function.
This is because the safe solution had a good evaluation value in this setting, and the projection to the safe region improved the evaluation value of the samples on the objective function, thereby accelerating the optimization process.

\begin{figure}[t]
    \centering
    \includegraphics[width=0.85\linewidth]{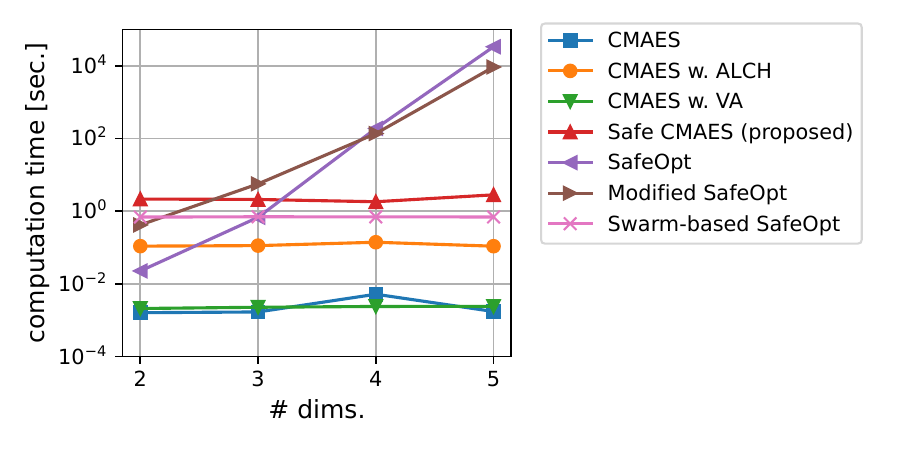}
    \caption{The computational time for performing five updates in each method. We plot the average time over three trials.}
    \label{fig:time}
\end{figure}
\begin{figure*}[t]
    \centering
    \includegraphics[width=0.86\linewidth]{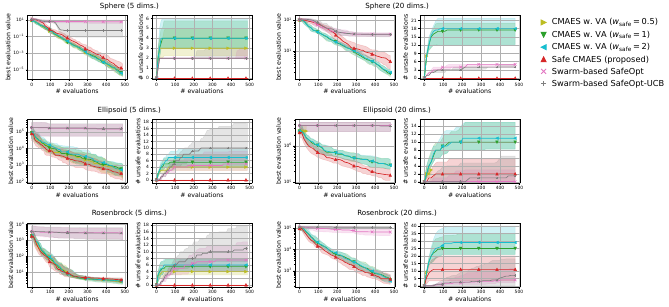}
    \caption{The performance comparison of the safe CMA-ES with the swarm-based SafeOpt and the CMA-ES with violation avoidance. We plot the medians and interquartile ranges of the best evaluation value and the number of \del{evaluations of unsafe solutions with safety function $s(\x) = f(\x)$}{}\nnew{unsafe evaluations}.}
    \label{fig:exp3}
\end{figure*}
%
%
%

\subsection{Result of Performance Evaluation Throughout Optimization Process}
\label{sec:exp_throughout}

We used the safety function and safety threshold to investigate the performance throughout the optimization process as
\begin{align}
    s(\x) = x_1 \qquad \text{and} \qquad h = 0  \enspace.
\end{align}
This safety function constrains the first element of the solution.
The optimization without evaluating unsafe solutions become challenging for the reversed ellipsoid function.
We set the total number of evaluations as $d \times 10^4$.

Figure~\ref{fig:exp2} shows the transitions of the best evaluation value satisfying the safety constraint and the number of unsafe evaluations.
It is noteworthy that the median numbers of the unsafe evaluations were zero \del{for all cases}{}\nnew{for the safe CMA-ES in all problems}.
The results on the sphere function show that the best evaluation value of the safe CMA-ES stagnated in the early phase of the optimization.
This occurred due to the overestimation of the Lipschitz constant of the safety function, and the modification of the initial step-size in~\eqref{eq:init:stepsize} set an initial value smaller than necessary.
On other functions, the safe CMA-ES required more evaluations to reduce the best evaluation value throughout the optimization compared to the comparative methods. 
The safe CMA-ES, especially on the 20-dimensional reversed ellipsoid, increased the number of evaluations by almost two times.
However, the safe CMA-ES completed the whole optimization process without evaluating unsafe solutions, in contrast to the comparative methods, which continued to increase unsafe solutions.
Additionally, the CMA-ES with violation avoidance failed to optimize the reversed ellipsoid and rosenbrock functions.
This results show the \new{safety and} efficiency of the safe CMA-ES on safe optimization.

\subsection{Result of Performance Comparison with Existing Methods}
\label{sec:exp_comparison}

Finally, we compared the safe CMA-ES with existing optimization methods for safe optimization.
Before evaluating the optimization performance, we compared the computational cost for updating.
Figure~\ref{fig:time} shows the computational time for performing five updates varying the number of dimensions as $d=2,3,4,5$.%
\footnote{
We measured the computational time using AMD EPYC 7763 (2.45GHz, 64 cores). We implemented the safe CMA-ES and violation avoidance using \texttt{NumPy~1.21.3}~\cite{numpy}, \texttt{SciPy~1.7.1}~\cite{scipy}, and \texttt{GpyTorch~1.10}~\cite{gpytorch}.
}
We observed a significant increase in the computational costs of SafeOpt and modified SafeOpt as the number of dimensions increased\nnew{, which made the comparison with the safe CMA-ES difficult}.
We consider that SafeOpt and modified SafeOpt used a grid space dividing the search space into even intervals, and the growing number of grid points resulted in an increased computational cost.%
\footnote{
We used a grid dividing each dimension into 20 points. 
We did not optimize the hyperparameters using the maximum likelihood method \del{to prevent early convergence}{}\nnew{because it deteriorated their performance in our preliminary experiment}.
}

Figure~\ref{fig:exp3} shows the comparison results between the safe CMA-ES, swarm-based SafeOpt, and violation avoidance using the safety constraint~\eqref{eq:exp1:constraint} as used in the first experiment.
We observed that swarm-based SafeOpt failed to reduce the best evaluation value on the ellipsoid and rosenbrock functions.
The reason for this failure is that the GPR could not accurately estimate the \new{original} landscape of those functions.
In contrast, because the safe CMA-ES uses the GPR on the composition function $s \circ \phi^{-1}$, the safe CMA-ES obtained a reliable estimation realizing the safe optimization efficiently.
Furthermore, more unsafe evaluations occurred in the violation avoidance than in the safe CMA-ES.
Additionally, in 20-dimensional problems, the violation avoidance with $w_\mathrm{safe} = 0.5$ terminated due to the inability to generate solutions whose nearest evaluated solutions were safe.
These results reveal that the continued superiority of the safe CMA-ES is not lost \del{by tuning the weights of the violation avoidance}{}\new{even when adjusting the weights of violation avoidance}.

\section{Conclusion}
We proposed the safe CMA-ES as an efficient optimization method tailored for safe optimization.
The safe CMA-ES estimates the Lipschitz constants of the transformed safety function using the distribution parameters. 
The estimation process uses GPR trained with $N_\mathrm{data}$ evaluated solutions and the maximum norm of the gradient $\|\nabla \mu(\z)\|$ on the mean of the posterior distribution.
Additionally, the safe CMA-ES constructs the safe region and projects the samples to the nearest points in the safe region to reduce the unsafe evaluations.
The safe CMA-ES also modifies the initial mean vector and initial step-size using the safe seeds.
The numerical simulation shows that the safe CMA-ES optimized the benchmark problems suppressing the unsafe evaluations \rev{although the rate of improvement in the best evaluation value was slower compared to other methods.}
As the safe CMA-ES assumes the existence of the Lipschitz constant of the safety functions, the algorithm improvement for safe optimization with discontinuous safety functions is considered as the future work.
Additionally, \rev{since we used the synthetic problems in our experiment, the evaluation of the safe CMA-ES in realistic problems is left as a future work.}

\begin{acks}
This work was partially supported by JSPS KAKENHI (JP23H00491, JP23H03466, JP23KJ0985), JST PRESTO (JPMJPR2133), and NEDO (JPNP18002, JPNP20006).
\end{acks}
\bibliographystyle{ACM-Reference-Format}
\bibliography{reference}


\begin{thebibliography}{25}


\ifx \showCODEN    \undefined \def \showCODEN     #1{\unskip}     \fi
\ifx \showDOI      \undefined \def \showDOI       #1{#1}\fi
\ifx \showISBNx    \undefined \def \showISBNx     #1{\unskip}     \fi
\ifx \showISBNxiii \undefined \def \showISBNxiii  #1{\unskip}     \fi
\ifx \showISSN     \undefined \def \showISSN      #1{\unskip}     \fi
\ifx \showLCCN     \undefined \def \showLCCN      #1{\unskip}     \fi
\ifx \shownote     \undefined \def \shownote      #1{#1}          \fi
\ifx \showarticletitle \undefined \def \showarticletitle #1{#1}   \fi
\ifx \showURL      \undefined \def \showURL       {\relax}        \fi
\providecommand\bibfield[2]{#2}
\providecommand\bibinfo[2]{#2}
\providecommand\natexlab[1]{#1}
\providecommand\showeprint[2][]{arXiv:#2}

\bibitem[\protect\citeauthoryear{Atamna, Auger, and Hansen}{Atamna
  et~al\mbox{.}}{2016}]%
        {alch:cmaes}
\bibfield{author}{\bibinfo{person}{Asma Atamna}, \bibinfo{person}{Anne Auger},
  {and} \bibinfo{person}{Nikolaus Hansen}.} \bibinfo{year}{2016}\natexlab{}.
\newblock \showarticletitle{Augmented Lagrangian Constraint Handling for
  {CMA-ES} --- Case of a Single Linear Constraint}. In
  \bibinfo{booktitle}{\emph{Parallel Problem Solving from Nature -- PPSN XIV}}.
  \bibinfo{publisher}{Springer International Publishing},
  \bibinfo{address}{Cham}, \bibinfo{pages}{181--191}.
\newblock
\showISBNx{978-3-319-45823-6}


\bibitem[\protect\citeauthoryear{Berkenkamp, Krause, and Schoellig}{Berkenkamp
  et~al\mbox{.}}{2023}]%
        {safeopt:constraint}
\bibfield{author}{\bibinfo{person}{Felix Berkenkamp}, \bibinfo{person}{Andreas
  Krause}, {and} \bibinfo{person}{Angela~P. Schoellig}.}
  \bibinfo{year}{2023}\natexlab{}.
\newblock \showarticletitle{{B}ayesian optimization with safety constraints:
  safe and automatic parameter tuning in robotics}.
\newblock \bibinfo{journal}{\emph{Machine Learning}} \bibinfo{volume}{112},
  \bibinfo{number}{10} (\bibinfo{year}{2023}), \bibinfo{pages}{3713--3747}.
\newblock
\showISBNx{1573-0565}
\urldef\tempurl%
\url{https://doi.org/10.1007/s10994-021-06019-1}
\showDOI{\tempurl}


\bibitem[\protect\citeauthoryear{Berkenkamp, Schoellig, and Krause}{Berkenkamp
  et~al\mbox{.}}{2016}]%
        {msafeopt}
\bibfield{author}{\bibinfo{person}{Felix Berkenkamp},
  \bibinfo{person}{Angela~P. Schoellig}, {and} \bibinfo{person}{Andreas
  Krause}.} \bibinfo{year}{2016}\natexlab{}.
\newblock \showarticletitle{Safe controller optimization for quadrotors with
  {G}aussian processes}. In \bibinfo{booktitle}{\emph{2016 IEEE International
  Conference on Robotics and Automation (ICRA)}}. \bibinfo{pages}{491--496}.
\newblock
\urldef\tempurl%
\url{https://doi.org/10.1109/ICRA.2016.7487170}
\showDOI{\tempurl}


\bibitem[\protect\citeauthoryear{Duivenvoorden, Berkenkamp, Carion, Krause, and
  Schoellig}{Duivenvoorden et~al\mbox{.}}{2017}]%
        {safeoptswarm}
\bibfield{author}{\bibinfo{person}{Rikky~R.P.R. Duivenvoorden},
  \bibinfo{person}{Felix Berkenkamp}, \bibinfo{person}{Nicolas Carion},
  \bibinfo{person}{Andreas Krause}, {and} \bibinfo{person}{Angela~P.
  Schoellig}.} \bibinfo{year}{2017}\natexlab{}.
\newblock \showarticletitle{Constrained {B}ayesian Optimization with Particle
  Swarms for Safe Adaptive Controller Tuning}.
\newblock \bibinfo{journal}{\emph{IFAC-PapersOnLine}} \bibinfo{volume}{50},
  \bibinfo{number}{1} (\bibinfo{year}{2017}), \bibinfo{pages}{11800--11807}.
\newblock
\showISSN{2405-8963}
\urldef\tempurl%
\url{https://doi.org/10.1016/j.ifacol.2017.08.1991}
\showDOI{\tempurl}
\newblock
\shownote{20th IFAC World Congress}.


\bibitem[\protect\citeauthoryear{Gardner, Pleiss, Bindel, Weinberger, and
  Wilson}{Gardner et~al\mbox{.}}{2018}]%
        {gpytorch}
\bibfield{author}{\bibinfo{person}{Jacob~R Gardner}, \bibinfo{person}{Geoff
  Pleiss}, \bibinfo{person}{David Bindel}, \bibinfo{person}{Kilian~Q
  Weinberger}, {and} \bibinfo{person}{Andrew~Gordon Wilson}.}
  \bibinfo{year}{2018}\natexlab{}.
\newblock \showarticletitle{{GPyTorch}: {B}lackbox Matrix-Matrix {G}aussian
  Process Inference with {GPU} Acceleration}. In
  \bibinfo{booktitle}{\emph{Advances in Neural Information Processing
  Systems}}.
\newblock


\bibitem[\protect\citeauthoryear{Gonzalez, Dai, Hennig, and Lawrence}{Gonzalez
  et~al\mbox{.}}{2016}]%
        {Gonzalez:2016}
\bibfield{author}{\bibinfo{person}{Javier Gonzalez}, \bibinfo{person}{Zhenwen
  Dai}, \bibinfo{person}{Philipp Hennig}, {and} \bibinfo{person}{Neil
  Lawrence}.} \bibinfo{year}{2016}\natexlab{}.
\newblock \showarticletitle{Batch {B}ayesian Optimization via Local
  Penalization}. In \bibinfo{booktitle}{\emph{Proceedings of the 19th
  International Conference on Artificial Intelligence and Statistics}},
  Vol.~\bibinfo{volume}{51}. \bibinfo{publisher}{PMLR},
  \bibinfo{pages}{648--657}.
\newblock


\bibitem[\protect\citeauthoryear{Hansen}{Hansen}{2016}]%
        {hansen:2017:arxiv}
\bibfield{author}{\bibinfo{person}{Nikolaus Hansen}.}
  \bibinfo{year}{2016}\natexlab{}.
\newblock \showarticletitle{The {CMA} Evolution Strategy: {A} Tutorial}.
\newblock \bibinfo{journal}{\emph{CoRR}}  \bibinfo{volume}{abs/1604.00772}
  (\bibinfo{year}{2016}).
\newblock
\showeprint[arXiv]{1604.00772}


\bibitem[\protect\citeauthoryear{Hansen, Akimoto, and Baudis}{Hansen
  et~al\mbox{.}}{2019}]%
        {hansen:pycma:2019}
\bibfield{author}{\bibinfo{person}{Nikolaus Hansen}, \bibinfo{person}{Youhei
  Akimoto}, {and} \bibinfo{person}{Petr Baudis}.}
  \bibinfo{year}{2019}\natexlab{}.
\newblock \bibinfo{booktitle}{\emph{{CMA-ES}/pycma on Github}}.
\newblock


\bibitem[\protect\citeauthoryear{Hansen and Auger}{Hansen and Auger}{2014}]%
        {hansen:2014:book}
\bibfield{author}{\bibinfo{person}{Nikolaus Hansen} {and} \bibinfo{person}{Anne
  Auger}.} \bibinfo{year}{2014}\natexlab{}.
\newblock \bibinfo{booktitle}{\emph{Principled Design of Continuous Stochastic
  Search: {F}rom Theory to Practice}}.
\newblock \bibinfo{publisher}{Springer Berlin Heidelberg},
  \bibinfo{address}{Berlin, Heidelberg}, \bibinfo{pages}{145--180}.
\newblock
\showISBNx{978-3-642-33206-7}
\urldef\tempurl%
\url{https://doi.org/10.1007/978-3-642-33206-7_8}
\showDOI{\tempurl}


\bibitem[\protect\citeauthoryear{Hansen and Ostermeier}{Hansen and
  Ostermeier}{1996}]%
        {hansen:1996:ec}
\bibfield{author}{\bibinfo{person}{Nikolaus Hansen} {and}
  \bibinfo{person}{Andreas Ostermeier}.} \bibinfo{year}{1996}\natexlab{}.
\newblock \showarticletitle{Adapting arbitrary normal mutation distributions in
  evolution strategies: {T}he covariance matrix adaptation}. In
  \bibinfo{booktitle}{\emph{Proceedings of IEEE International Conference on
  Evolutionary Computation}}. \bibinfo{pages}{312--317}.
\newblock
\urldef\tempurl%
\url{https://doi.org/10.1109/ICEC.1996.542381}
\showDOI{\tempurl}


\bibitem[\protect\citeauthoryear{Harris, Millman, van~der Walt, Gommers,
  Virtanen, Cournapeau, Wieser, Taylor, Berg, Smith, Kern, Picus, Hoyer, van
  Kerkwijk, Brett, Haldane, del R{\'{i}}o, Wiebe, Peterson,
  G{\'{e}}rard-Marchant, Sheppard, Reddy, Weckesser, Abbasi, Gohlke, and
  Oliphant}{Harris et~al\mbox{.}}{2020}]%
        {numpy}
\bibfield{author}{\bibinfo{person}{Charles~R. Harris},
  \bibinfo{person}{K.~Jarrod Millman}, \bibinfo{person}{St{'{e}}fan~J. van~der
  Walt}, \bibinfo{person}{Ralf Gommers}, \bibinfo{person}{Pauli Virtanen},
  \bibinfo{person}{David Cournapeau}, \bibinfo{person}{Eric Wieser},
  \bibinfo{person}{Julian Taylor}, \bibinfo{person}{Sebastian Berg},
  \bibinfo{person}{Nathaniel~J. Smith}, \bibinfo{person}{Robert Kern},
  \bibinfo{person}{Matti Picus}, \bibinfo{person}{Stephan Hoyer},
  \bibinfo{person}{Marten~H. van Kerkwijk}, \bibinfo{person}{Matthew Brett},
  \bibinfo{person}{Allan Haldane}, \bibinfo{person}{Jaime~Fern{\'{a}}ndez del
  R{\'{i}}o}, \bibinfo{person}{Mark Wiebe}, \bibinfo{person}{Pearu Peterson},
  \bibinfo{person}{Pierre G{\'{e}}rard-Marchant}, \bibinfo{person}{Kevin
  Sheppard}, \bibinfo{person}{Tyler Reddy}, \bibinfo{person}{Warren Weckesser},
  \bibinfo{person}{Hameer Abbasi}, \bibinfo{person}{Christoph Gohlke}, {and}
  \bibinfo{person}{Travis~E. Oliphant}.} \bibinfo{year}{2020}\natexlab{}.
\newblock \showarticletitle{Array programming with {NumPy}}.
\newblock \bibinfo{journal}{\emph{Nature}} \bibinfo{volume}{585},
  \bibinfo{number}{7825} (\bibinfo{year}{2020}), \bibinfo{pages}{357--362}.
\newblock
\urldef\tempurl%
\url{https://doi.org/10.1038/s41586-020-2649-2}
\showDOI{\tempurl}


\bibitem[\protect\citeauthoryear{Hutter, Hoos, and Leyton-Brown}{Hutter
  et~al\mbox{.}}{2013}]%
        {smac:bbob}
\bibfield{author}{\bibinfo{person}{Frank Hutter}, \bibinfo{person}{Holger
  Hoos}, {and} \bibinfo{person}{Kevin Leyton-Brown}.}
  \bibinfo{year}{2013}\natexlab{}.
\newblock \showarticletitle{An evaluation of sequential model-based
  optimization for expensive blackbox functions}. In
  \bibinfo{booktitle}{\emph{Proceedings of the 15th Annual Conference Companion
  on Genetic and Evolutionary Computation}}. \bibinfo{publisher}{Association
  for Computing Machinery}, \bibinfo{address}{New York, NY, USA},
  \bibinfo{pages}{1209--1216}.
\newblock
\showISBNx{9781450319645}
\urldef\tempurl%
\url{https://doi.org/10.1145/2464576.2501592}
\showDOI{\tempurl}


\bibitem[\protect\citeauthoryear{Kaji, Ikeda, and Kita}{Kaji
  et~al\mbox{.}}{2009}]%
        {Kaji:2009}
\bibfield{author}{\bibinfo{person}{Hirotaka Kaji}, \bibinfo{person}{Kokolo
  Ikeda}, {and} \bibinfo{person}{Hajime Kita}.}
  \bibinfo{year}{2009}\natexlab{}.
\newblock \showarticletitle{Avoidance of constraint violation for
  experiment-based evolutionary multi-objective optimization}. In
  \bibinfo{booktitle}{\emph{2009 IEEE Congress on Evolutionary Computation}}.
  \bibinfo{pages}{2756--2763}.
\newblock
\urldef\tempurl%
\url{https://doi.org/10.1109/CEC.2009.4983288}
\showDOI{\tempurl}


\bibitem[\protect\citeauthoryear{Kim, Allmendinger, and
  L\'{o}pez-Ib\'{a}\~{n}ez}{Kim et~al\mbox{.}}{2022}]%
        {Kim:2022}
\bibfield{author}{\bibinfo{person}{Youngmin Kim}, \bibinfo{person}{Richard
  Allmendinger}, {and} \bibinfo{person}{Manuel L\'{o}pez-Ib\'{a}\~{n}ez}.}
  \bibinfo{year}{2022}\natexlab{}.
\newblock \showarticletitle{Are Evolutionary Algorithms Safe Optimizers?}. In
  \bibinfo{booktitle}{\emph{Proceedings of the Genetic and Evolutionary
  Computation Conference}}. \bibinfo{publisher}{Association for Computing
  Machinery}, \bibinfo{pages}{814--822}.
\newblock
\showISBNx{9781450392372}
\urldef\tempurl%
\url{https://doi.org/10.1145/3512290.3528818}
\showDOI{\tempurl}


\bibitem[\protect\citeauthoryear{Kim, Allmendinger, and
  L{\'o}pez-Ib{\'a}{\~{n}}ez}{Kim et~al\mbox{.}}{2021}]%
        {survey}
\bibfield{author}{\bibinfo{person}{Youngmin Kim}, \bibinfo{person}{Richard
  Allmendinger}, {and} \bibinfo{person}{Manuel L{\'o}pez-Ib{\'a}{\~{n}}ez}.}
  \bibinfo{year}{2021}\natexlab{}.
\newblock \showarticletitle{Safe Learning and Optimization Techniques:
  {T}owards a Survey of the State of the Art}. In
  \bibinfo{booktitle}{\emph{Trustworthy AI - Integrating Learning, Optimization
  and Reasoning}}. \bibinfo{publisher}{Springer International Publishing},
  \bibinfo{address}{Cham}, \bibinfo{pages}{123--139}.
\newblock
\showISBNx{978-3-030-73959-1}


\bibitem[\protect\citeauthoryear{Liu and Nocedal}{Liu and Nocedal}{1989}]%
        {lbfgs}
\bibfield{author}{\bibinfo{person}{Dong~C Liu} {and} \bibinfo{person}{Jorge
  Nocedal}.} \bibinfo{year}{1989}\natexlab{}.
\newblock \showarticletitle{On the limited memory {BFGS} method for large scale
  optimization}.
\newblock \bibinfo{journal}{\emph{Mathematical Programming}}
  \bibinfo{volume}{45}, \bibinfo{number}{1} (\bibinfo{year}{1989}),
  \bibinfo{pages}{503--528}.
\newblock
\showISSN{1436-4646}
\urldef\tempurl%
\url{https://doi.org/10.1007/BF01589116}
\showDOI{\tempurl}


\bibitem[\protect\citeauthoryear{Louis, Ugalde, Gauthier, Adenis, Tourki, and
  Huneker}{Louis et~al\mbox{.}}{2022}]%
        {Louis:2022}
\bibfield{author}{\bibinfo{person}{Maxime Louis},
  \bibinfo{person}{Hector~Romero Ugalde}, \bibinfo{person}{Pierre Gauthier},
  \bibinfo{person}{Alice Adenis}, \bibinfo{person}{Yousra Tourki}, {and}
  \bibinfo{person}{Erik Huneker}.} \bibinfo{year}{2022}\natexlab{}.
\newblock \showarticletitle{Safe Reinforcement Learning for Automatic Insulin
  Delivery in Type I Diabetes}. In \bibinfo{booktitle}{\emph{{Reinforcement
  Learning for Real Life Workshop, NeurIPS 2022}}}.
\newblock


\bibitem[\protect\citeauthoryear{Modugno, Chervet, Oriolo, and Ivaldi}{Modugno
  et~al\mbox{.}}{2016}]%
        {Modugno:2016}
\bibfield{author}{\bibinfo{person}{Valerio Modugno}, \bibinfo{person}{Ugo
  Chervet}, \bibinfo{person}{Giuseppe Oriolo}, {and} \bibinfo{person}{Serena
  Ivaldi}.} \bibinfo{year}{2016}\natexlab{}.
\newblock \showarticletitle{Learning soft task priorities for safe control of
  humanoid robots with constrained stochastic optimization}. In
  \bibinfo{booktitle}{\emph{2016 IEEE-RAS 16th International Conference on
  Humanoid Robots (Humanoids)}}. \bibinfo{pages}{101--108}.
\newblock
\urldef\tempurl%
\url{https://doi.org/10.1109/HUMANOIDS.2016.7803261}
\showDOI{\tempurl}


\bibitem[\protect\citeauthoryear{Nomura and Shibata}{Nomura and
  Shibata}{2024}]%
        {nomura2024cmaes}
\bibfield{author}{\bibinfo{person}{Masahiro Nomura} {and}
  \bibinfo{person}{Masashi Shibata}.} \bibinfo{year}{2024}\natexlab{}.
\newblock \showarticletitle{cmaes : A Simple yet Practical Python Library for
  CMA-ES}.
\newblock \bibinfo{journal}{\emph{arXiv preprint arXiv:2402.01373}}
  (\bibinfo{year}{2024}).
\newblock


\bibitem[\protect\citeauthoryear{Rasmussen, Williams, et~al\mbox{.}}{Rasmussen
  et~al\mbox{.}}{2006}]%
        {rasmussen2006gaussian}
\bibfield{author}{\bibinfo{person}{Carl~Edward Rasmussen},
  \bibinfo{person}{Christopher~KI Williams}, {et~al\mbox{.}}}
  \bibinfo{year}{2006}\natexlab{}.
\newblock \bibinfo{booktitle}{\emph{Gaussian Processes for Machine Learning}}.
  Vol.~\bibinfo{volume}{1}.
\newblock \bibinfo{publisher}{Springer}.
\newblock


\bibitem[\protect\citeauthoryear{Serkan~Kiranyaz}{Serkan~Kiranyaz}{2014}]%
        {pso}
\bibfield{author}{\bibinfo{person}{Moncef~Gabbouj Serkan~Kiranyaz,
  Turker~Ince}.} \bibinfo{year}{2014}\natexlab{}.
\newblock \bibinfo{booktitle}{\emph{Multidimensional Particle Swarm
  Optimization for Machine Learning and Pattern Recognition}
  (\bibinfo{edition}{first} ed.)}.
\newblock \bibinfo{publisher}{Springer}.
\newblock
\urldef\tempurl%
\url{https://doi.org/10.1007/978-3-642-37846-1}
\showDOI{\tempurl}


\bibitem[\protect\citeauthoryear{Sui, Gotovos, Burdick, and Krause}{Sui
  et~al\mbox{.}}{2015}]%
        {safeopt}
\bibfield{author}{\bibinfo{person}{Yanan Sui}, \bibinfo{person}{Alkis Gotovos},
  \bibinfo{person}{Joel Burdick}, {and} \bibinfo{person}{Andreas Krause}.}
  \bibinfo{year}{2015}\natexlab{}.
\newblock \showarticletitle{Safe Exploration for Optimization with {G}aussian
  Processes}. In \bibinfo{booktitle}{\emph{Proceedings of the 32nd
  International Conference on Machine Learning}}
  \emph{(\bibinfo{series}{Proceedings of Machine Learning Research},
  Vol.~\bibinfo{volume}{37})}. \bibinfo{publisher}{PMLR},
  \bibinfo{pages}{997--1005}.
\newblock


\bibitem[\protect\citeauthoryear{Toal and Arnold}{Toal and Arnold}{2020}]%
        {gpcmaes}
\bibfield{author}{\bibinfo{person}{Lauchlan Toal} {and}
  \bibinfo{person}{Dirk~V. Arnold}.} \bibinfo{year}{2020}\natexlab{}.
\newblock \showarticletitle{Simple Surrogate Model Assisted Optimization with
  Covariance Matrix Adaptation}. In \bibinfo{booktitle}{\emph{Parallel Problem
  Solving from Nature -- PPSN XVI}}. \bibinfo{publisher}{Springer International
  Publishing}, \bibinfo{pages}{184--197}.
\newblock
\showISBNx{978-3-030-58112-1}


\bibitem[\protect\citeauthoryear{Virtanen, Gommers, Oliphant, Haberland, Reddy,
  Cournapeau, Burovski, Peterson, Weckesser, Bright, {van der Walt}, Brett,
  Wilson, Millman, Mayorov, Nelson, Jones, Kern, Larson, Carey, Polat, Feng,
  Moore, {VanderPlas}, Laxalde, Perktold, Cimrman, Henriksen, Quintero, Harris,
  Archibald, Ribeiro, Pedregosa, {van Mulbregt}, and {SciPy 1.0
  Contributors}}{Virtanen et~al\mbox{.}}{2020}]%
        {scipy}
\bibfield{author}{\bibinfo{person}{Pauli Virtanen}, \bibinfo{person}{Ralf
  Gommers}, \bibinfo{person}{Travis~E. Oliphant}, \bibinfo{person}{Matt
  Haberland}, \bibinfo{person}{Tyler Reddy}, \bibinfo{person}{David
  Cournapeau}, \bibinfo{person}{Evgeni Burovski}, \bibinfo{person}{Pearu
  Peterson}, \bibinfo{person}{Warren Weckesser}, \bibinfo{person}{Jonathan
  Bright}, \bibinfo{person}{St{\'e}fan~J. {van der Walt}},
  \bibinfo{person}{Matthew Brett}, \bibinfo{person}{Joshua Wilson},
  \bibinfo{person}{K.~Jarrod Millman}, \bibinfo{person}{Nikolay Mayorov},
  \bibinfo{person}{Andrew R.~J. Nelson}, \bibinfo{person}{Eric Jones},
  \bibinfo{person}{Robert Kern}, \bibinfo{person}{Eric Larson},
  \bibinfo{person}{C~J Carey}, \bibinfo{person}{{\.I}lhan Polat},
  \bibinfo{person}{Yu Feng}, \bibinfo{person}{Eric~W. Moore},
  \bibinfo{person}{Jake {VanderPlas}}, \bibinfo{person}{Denis Laxalde},
  \bibinfo{person}{Josef Perktold}, \bibinfo{person}{Robert Cimrman},
  \bibinfo{person}{Ian Henriksen}, \bibinfo{person}{E.~A. Quintero},
  \bibinfo{person}{Charles~R. Harris}, \bibinfo{person}{Anne~M. Archibald},
  \bibinfo{person}{Ant{\^o}nio~H. Ribeiro}, \bibinfo{person}{Fabian Pedregosa},
  \bibinfo{person}{Paul {van Mulbregt}}, {and} \bibinfo{person}{{SciPy 1.0
  Contributors}}.} \bibinfo{year}{2020}\natexlab{}.
\newblock \showarticletitle{{{SciPy} 1.0: {F}undamental Algorithms for
  Scientific Computing in {P}ython}}.
\newblock \bibinfo{journal}{\emph{Nature Methods}}  \bibinfo{volume}{17}
  (\bibinfo{year}{2020}), \bibinfo{pages}{261--272}.
\newblock
\urldef\tempurl%
\url{https://doi.org/10.1038/s41592-019-0686-2}
\showDOI{\tempurl}


\bibitem[\protect\citeauthoryear{Yamada, Uchida, Saito, and Shirakawa}{Yamada
  et~al\mbox{.}}{2023}]%
        {sgpcmaes}
\bibfield{author}{\bibinfo{person}{Yutaro Yamada}, \bibinfo{person}{Kento
  Uchida}, \bibinfo{person}{Shota Saito}, {and} \bibinfo{person}{Shinichi
  Shirakawa}.} \bibinfo{year}{2023}\natexlab{}.
\newblock \showarticletitle{Surrogate-Assisted (1+1)-CMA-ES with Switching
  Mechanism of Utility Functions}. In \bibinfo{booktitle}{\emph{Applications of
  Evolutionary Computation}}. \bibinfo{publisher}{Springer Nature Switzerland},
  \bibinfo{pages}{798--814}.
\newblock
\showISBNx{978-3-031-30229-9}


\end{thebibliography}

\clearpage

\appendix

\section{Investigation of Hyperparameter Sensitivity}
We investigated the sensitivities of hyperparameters $\alpha$, $\zeta_\mathrm{init}$, and $T_\mathrm{data}$ of the safe CMA-ES. 
The effect of those hyperparameters are summarized as
\begin{itemize}
\setlength{\leftskip}{-0.35cm}
    \item The hyperparameter $\alpha$ controls the rates of increase and decrease for the coefficient $\rho_j^{(t+1)}$ for the estimated Lipschitz constant when the unsafe solutions are evaluated.
    \item The hyperparameter $\zeta_\mathrm{init}$ controls the coefficient $\tau^{(t+1)}$ for the estimated Lipschitz constant when the number of evaluated solutions used for the Gaussian process regression is small.
    \item The hyperparameter $T_\mathrm{data}$ controls the number $N_\mathrm{data}$ of solutions to construct the safe region and the number $N_\mathrm{data}$ of training data for Gaussian process regression.
\end{itemize}

We used the safety constraint used in the first and third experiments in the paper as
\begin{align}
    s(\x) = f(\x) \quad \text{and} \quad h = q(f, \X, 0.5) \enspace.
\end{align}
We ran the safe CMA-ES changing a single hyperparameter and remaining the other hyperparameters to their recommended settings.
We ran 50 independent trials for each setting.

\subsection{Result of Sensitivity Experiment of $\alpha$}
\begin{figure*}[t]
    \centering
    \begin{minipage}[b]{0.45\linewidth}
    \centering
    \includegraphics[width=\linewidth]{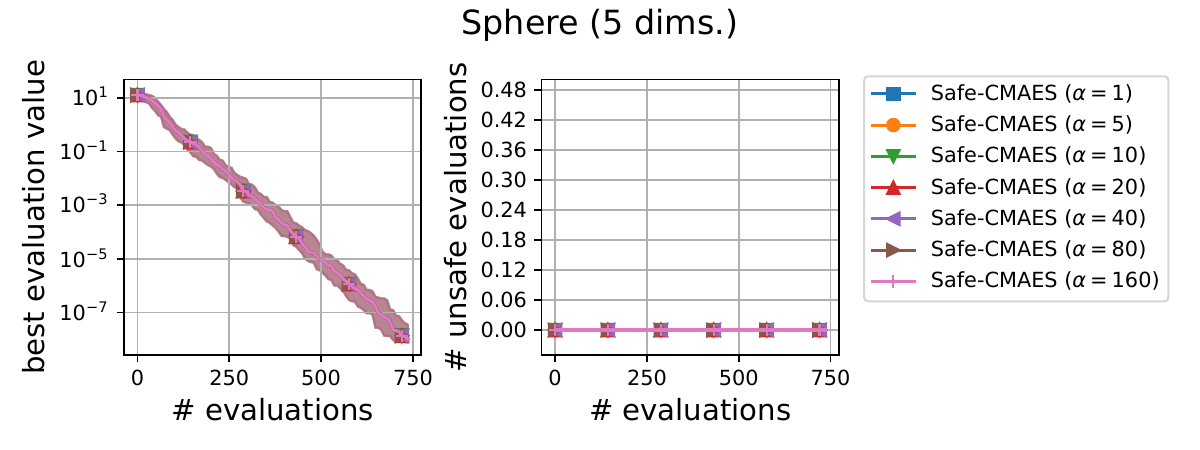}
    \end{minipage}
    \begin{minipage}[b]{0.45\linewidth}
    \centering
    \includegraphics[width=\linewidth]{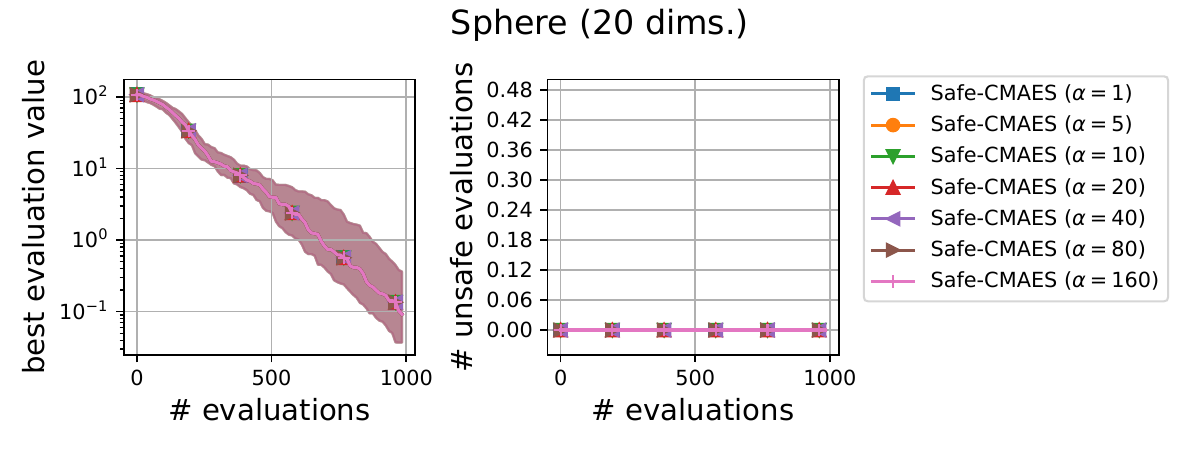}
    \end{minipage} 
    \begin{minipage}[b]{0.45\linewidth}
    \centering
    \includegraphics[width=\linewidth]{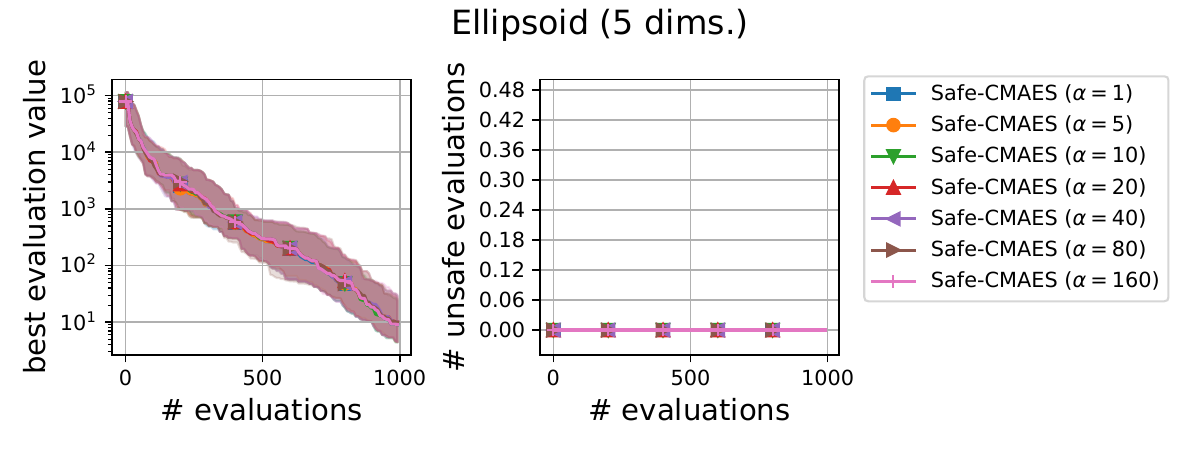}
    \end{minipage}
    \begin{minipage}[b]{0.45\linewidth}
    \centering
    \includegraphics[width=\linewidth]{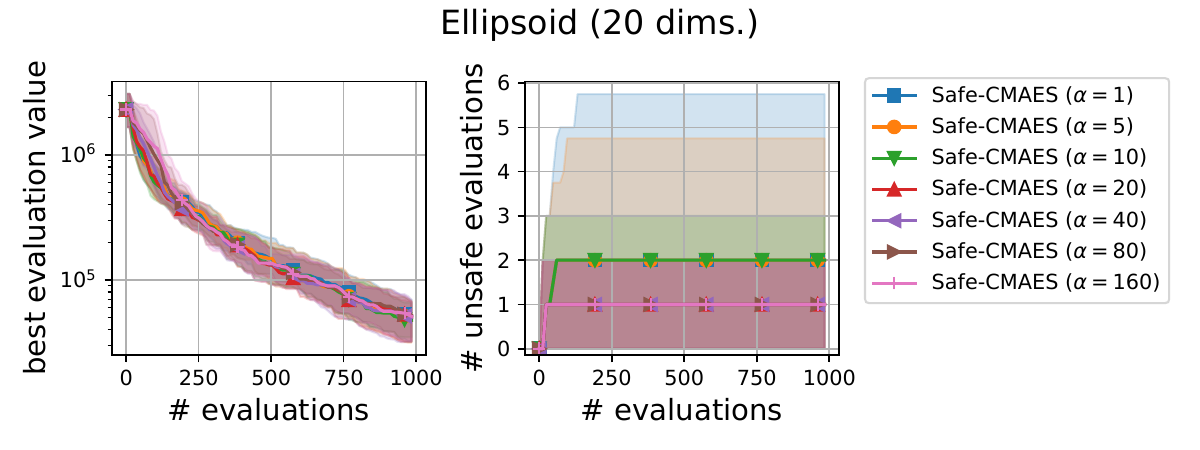}
    \end{minipage}
    \begin{minipage}[b]{0.45\linewidth}
    \centering
    \includegraphics[width=\linewidth]{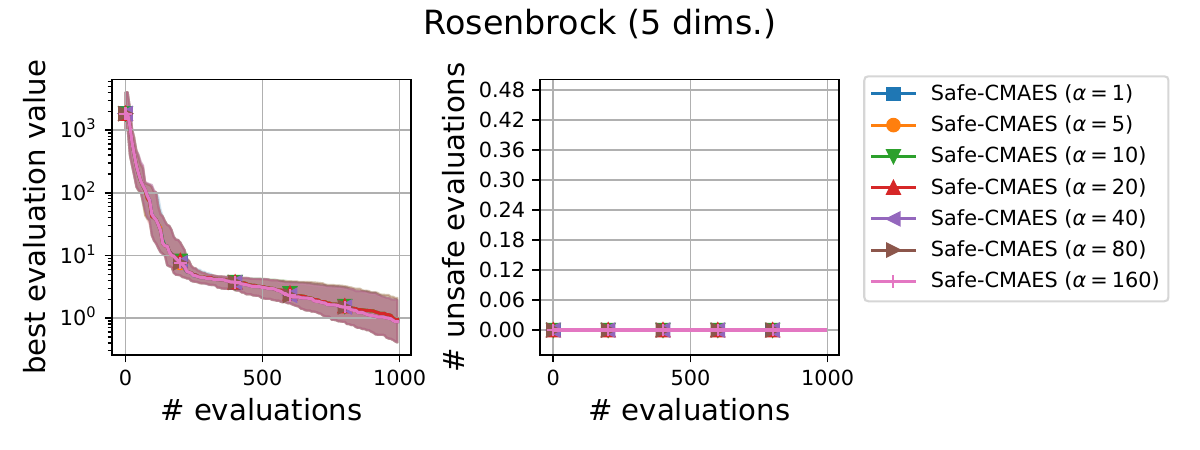}
    \end{minipage}
    \begin{minipage}[b]{0.45\linewidth}
    \centering
    \includegraphics[width=\linewidth]{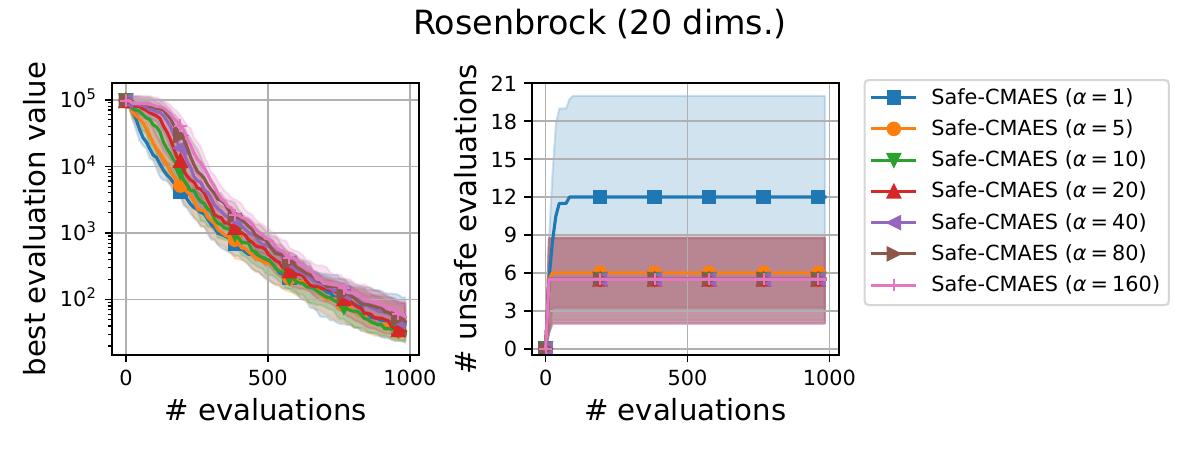}
    \end{minipage}
    \caption{Result of Sensitivity Experiment of $\alpha$. We plot the medians and interquartile ranges of the best evaluation value and the number of evaluations of unsafe solutions.}
    \label{fig:alpha}
\end{figure*}

We varied the setting of $\alpha$ as $\alpha = 1, 5, 10, 20, 40, 80, 160$.
Figure~\ref{fig:alpha} shows the transitions of the best evaluation value and the number of unsafe evaluations.
We note that no unsafe evaluation occured in 5-dimensional functions and the 20-dimensional sphere function.
Since the hyperparameter $\alpha$ affects the dynamics only when an unsafe evaluation occurs, the dynamics of the safe CMA-ES were not changed in those cases.
For the 20-dimensional ellipsoid function, the unsafe evaluations tended to be reduced as the hyperparameter $\alpha$ increased, while the best evaluation value slightly stagnated in the first few updates.
On the 20-dimensional rosenbrock function, a similar stagnation of the best evaluation value was observed with large $\alpha$, while the number of unsafe evaluations remained unchanged except for the case $\alpha = 1$, i.e., the case without the adaptation of the coefficient $\rho_j^{(t+1)}$.
Overall, the recommended setting $\alpha = 10$ seems to be a reasonable choice.

\subsection{Result of Sensitivity Experiment of $\zeta_\mathrm{init}$}
\begin{figure*}[t]
    \centering
    \begin{minipage}[b]{0.45\linewidth}
    \centering
    \includegraphics[width=\linewidth]{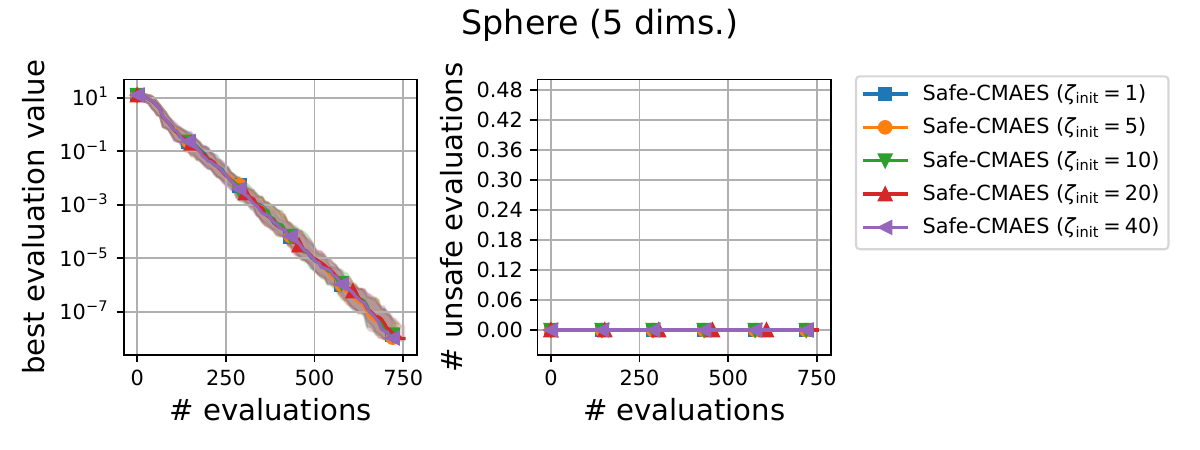}
    \end{minipage}
    \begin{minipage}[b]{0.45\linewidth}
    \centering
    \includegraphics[width=\linewidth]{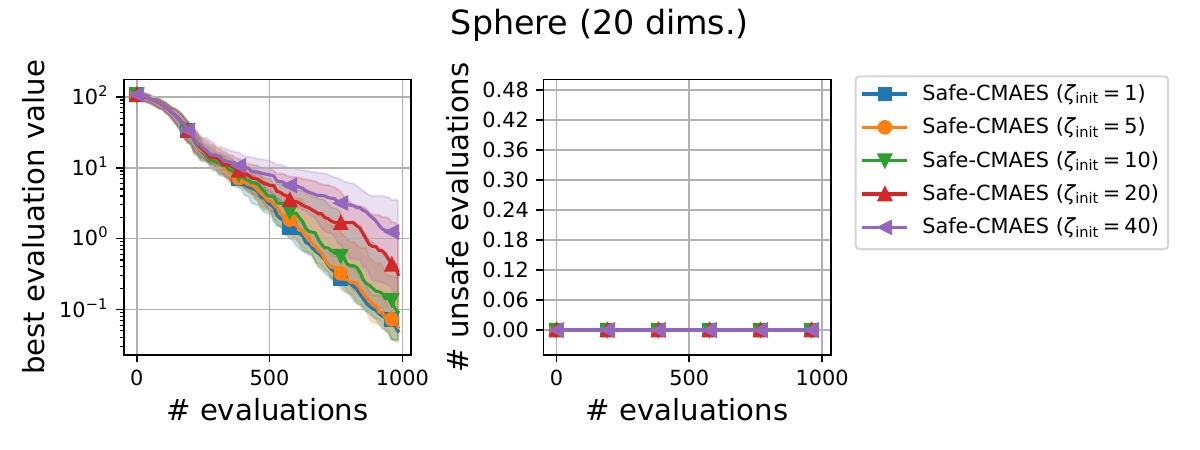}
    \end{minipage} 
    \begin{minipage}[b]{0.45\linewidth}
    \centering
    \includegraphics[width=\linewidth]{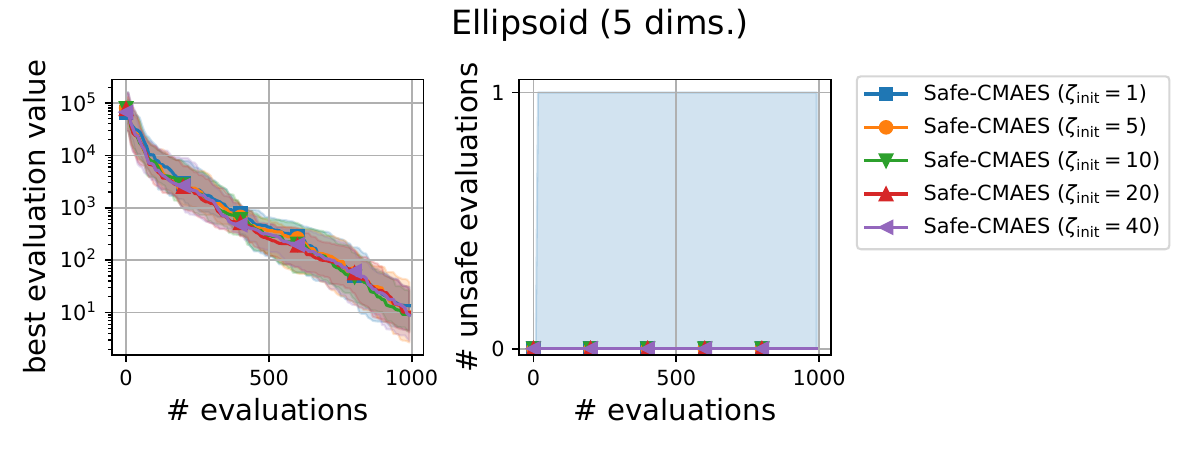}
    \end{minipage}
    \begin{minipage}[b]{0.45\linewidth}
    \centering
    \includegraphics[width=\linewidth]{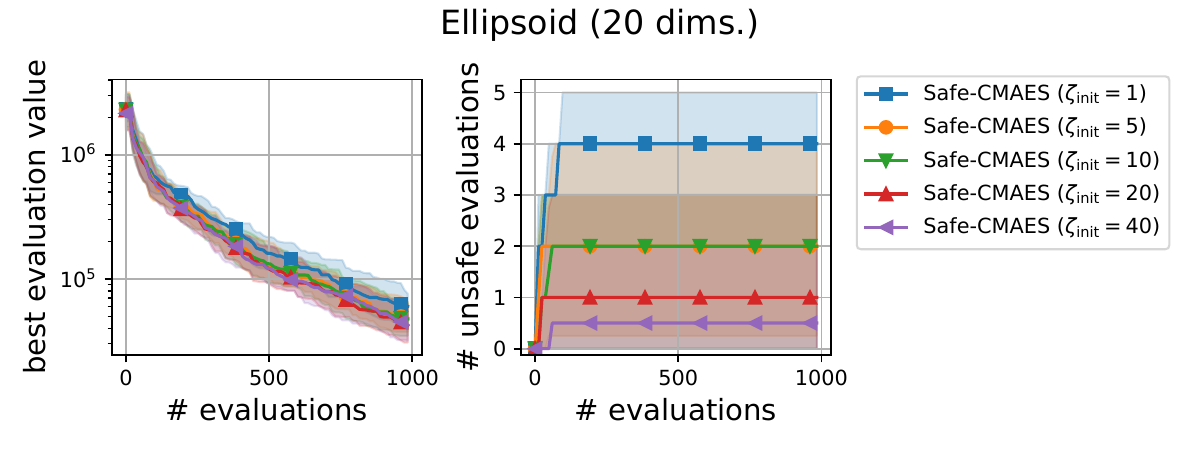}
    \end{minipage}
    \begin{minipage}[b]{0.45\linewidth}
    \centering
    \includegraphics[width=\linewidth]{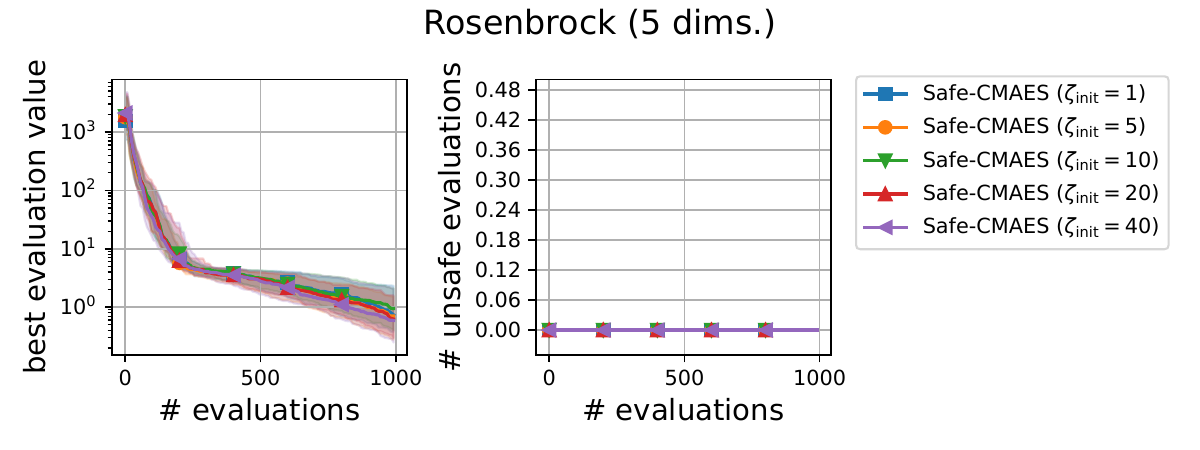}
    \end{minipage}
    \begin{minipage}[b]{0.45\linewidth}
    \centering
    \includegraphics[width=\linewidth]{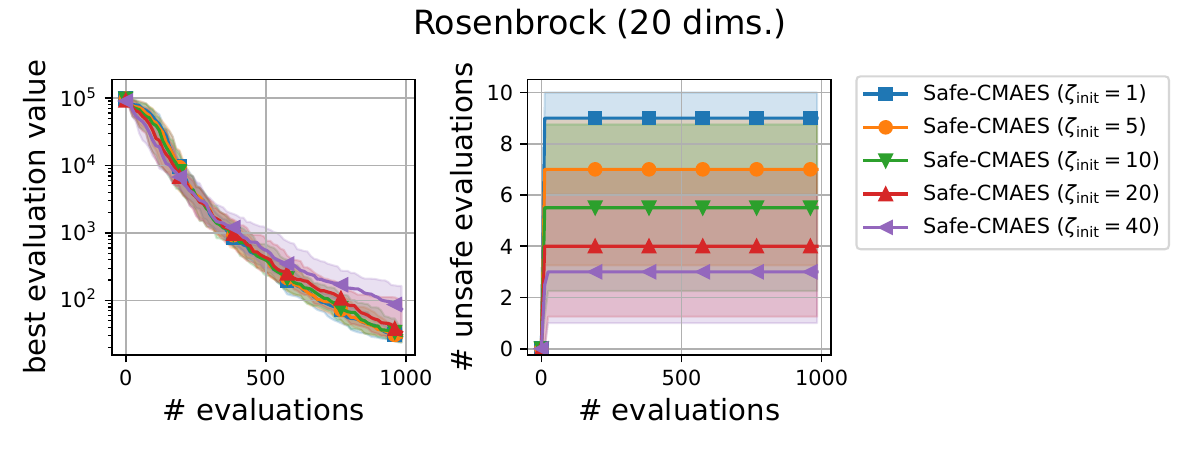}
    \end{minipage}
    \caption{Result of Sensitivity Experiment of $\zeta_\mathrm{init}$. We plot the medians and interquartile ranges of the best evaluation value and the number of evaluations of unsafe solutions.}
    \label{fig:zeta}
\end{figure*}

We varied the setting of $\zeta_\mathrm{init}$ as $\zeta_\mathrm{init} = 1, 5, 10, 20, 40$.
Figure~\ref{fig:zeta} shows the transitions of the best evaluation value and the number of unsafe evaluations.
When $\zeta_\mathrm{init} = 1$, indicating that the adaptation of coefficient $\tau^{(t+1)}$ is not applied, the unsafe evaluations were occasionally occurred on the 5-dimensional ellipsoid.
On the 20-dimensional sphere function, a larger setting of $\zeta_\mathrm{init}$ delayed the decrease of the best evaluation value, while unsafe evaluation did not occur in any of the settings.
In contrast, on the 20-dimensional ellipsoid function, a larger setting of $\zeta_\mathrm{init}$ reduced the number of unsafe evaluations while maintaining the decrease rate in the best evaluation value.
We consider the performance of the safe CMA-ES to be a somewhat sensitive to $\zeta_\mathrm{init}$, and the suitable setting of $\zeta_\mathrm{init}$ depends on the problem.

\subsection{Result of Sensitivity Experiment of $T_\mathrm{data}$}
\begin{figure*}[t]
    \centering
    \begin{minipage}[b]{0.45\linewidth}
    \centering
    \includegraphics[width=\linewidth]{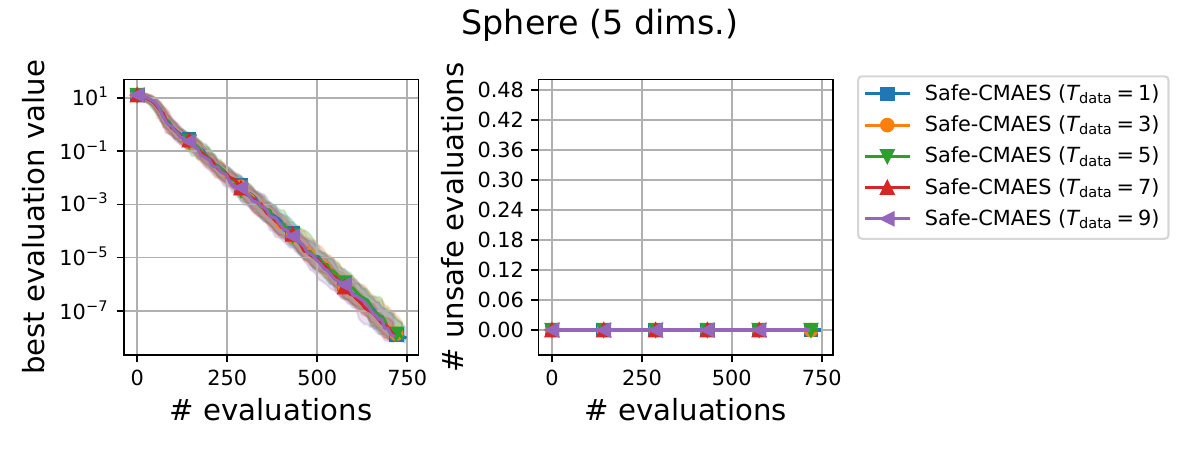}
    \end{minipage}
    \begin{minipage}[b]{0.45\linewidth}
    \centering
    \includegraphics[width=\linewidth]{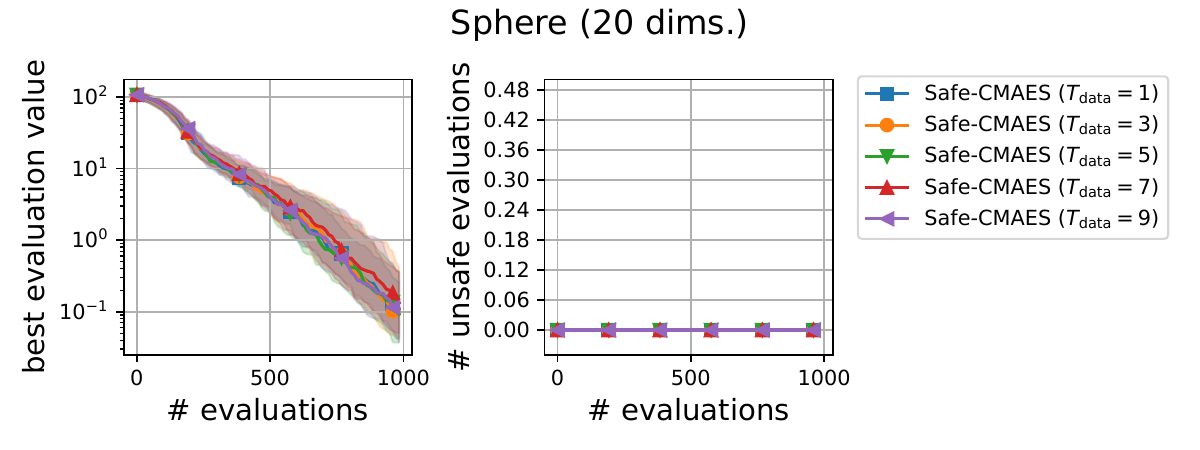}
    \end{minipage} 
    \begin{minipage}[b]{0.45\linewidth}
    \centering
    \includegraphics[width=\linewidth]{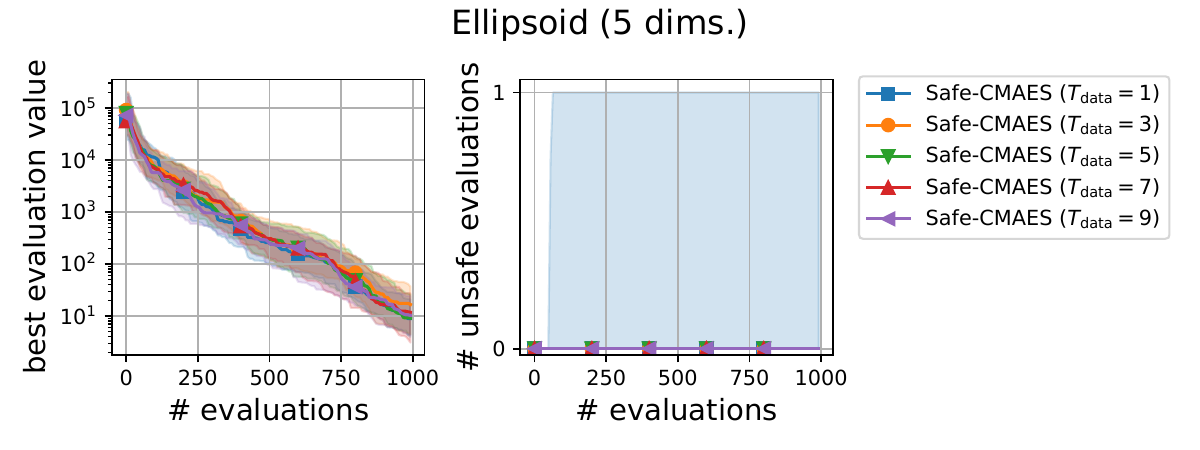}
    \end{minipage}
    \begin{minipage}[b]{0.45\linewidth}
    \centering
    \includegraphics[width=\linewidth]{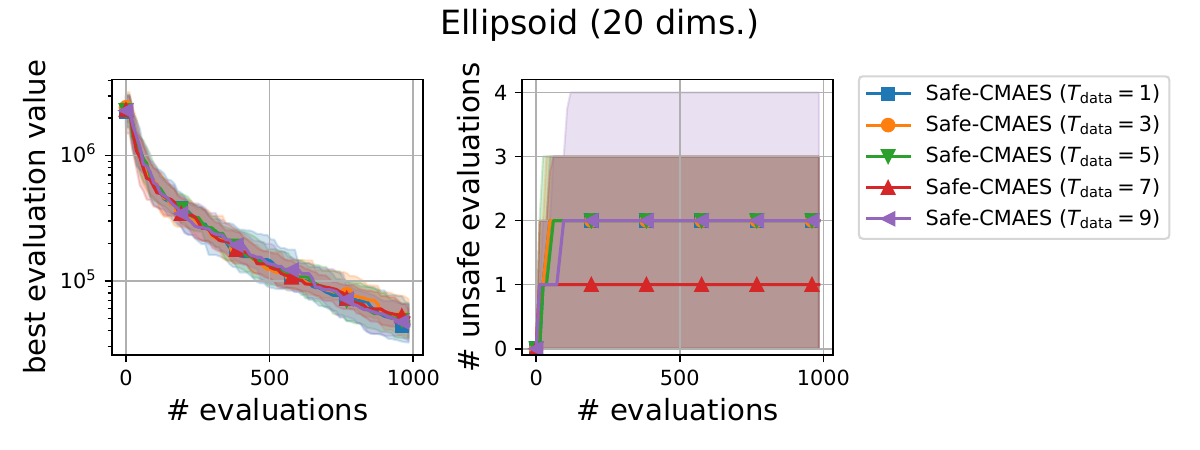}
    \end{minipage}
    \begin{minipage}[b]{0.45\linewidth}
    \centering
    \includegraphics[width=\linewidth]{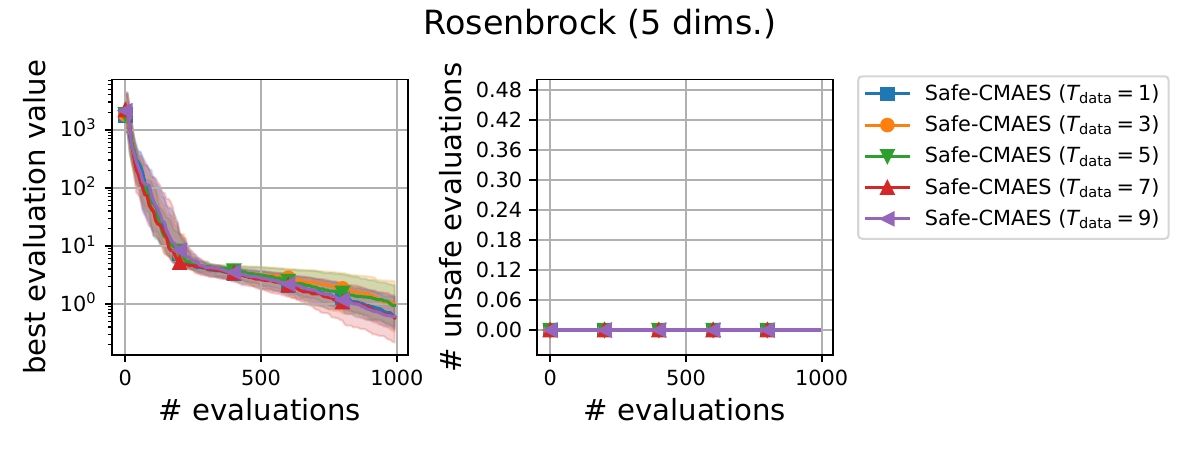}
    \end{minipage}
    \begin{minipage}[b]{0.45\linewidth}
    \centering
    \includegraphics[width=\linewidth]{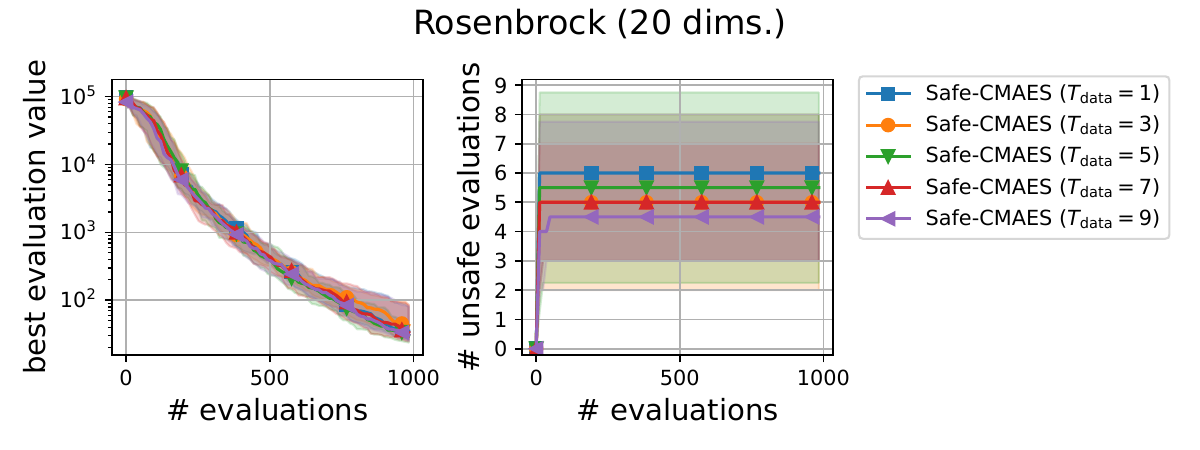}
    \end{minipage}
    \caption{Result of Sensitivity Experiment of $T_\mathrm{data}$. We plot the medians and interquartile ranges of the best evaluation value and the number of evaluations of unsafe solutions.}
    \label{fig:Tdata}
\end{figure*}

We varied the setting of $T_\mathrm{data}$ as $T_\mathrm{data} = 1, 3, 5, 7, 9$.
Figure~\ref{fig:Tdata} shows the transitions of the best evaluation value and the number of unsafe evaluations.
The setting of $T_\mathrm{data}$ did not significantly change the dynamics of the safe CMA-ES significantly on most of the problems.
When $T_\mathrm{data} = 1$, the unsafe evaluation occurred in the 5-dimensional ellipsoid function.
On the 20-dimensional rosenbrock function, a lager $T_\mathrm{data}$ reduced the unsafe evaluations.
Overall, the recommended setting $T_\mathrm{data} = 5$ performed well across all problems.


\end{document}


\title{CMA-ES for Safe Optimization: Supplementary Material}


\author{Kento Uchida}
\email{uchida-kento-fz@ynu.ac.jp}
\orcid{0000-0002-4179-6020}
\affiliation{%
  \institution{Yokohama National University}
  \city{Yokohama}
  \state{Kanagawa}
  \country{Japan}
  \postcode{240-8501}
}

\author{Ryoki Hamano}
\email{hamano-ryoki-pd@ynu.jp}
\orcid{0000-0002-4425-1683}
\affiliation{%
  \institution{Yokohama National University}
  \city{Yokohama}
  \state{Kanagawa}
  \country{Japan}
  \postcode{240-8501}
}

\author{Masahiro Nomura}
\email{nomura\_masahiro@cyberagent.co.jp}
\orcid{0000-0002-4945-5984}
\affiliation{%
  \institution{CyberAgent, Inc.}
  \city{Shibuya}
  \state{Tokyo}
  \country{Japan}
  \postcode{150-0042}
}

\author{Shota Saito}
\email{saito-shota-bt@ynu.jp}
\orcid{0000-0002-9863-6765}
\affiliation{%
  \institution{Yokohama National University \and SKILLUP NeXt Ltd.}
  \city{Yokohama}
  \state{Kanagawa}
  \country{Japan}
  \postcode{240-8501}
}

\author{Shinichi Shirakawa}
\email{shirakawa-shinichi-bg@ynu.ac.jp}
\orcid{0000-0002-4659-6108}
\affiliation{%
  \institution{Yokohama National University}
  \city{Yokohama}
  \state{Kanagawa}
  \country{Japan}
  \postcode{240-8501}
}

\renewcommand{\shortauthors}{K. Uchida et al.}

\maketitle

\appendix

\begin{figure*}[t]
    \centering
    \includegraphics[width=0.9\linewidth]{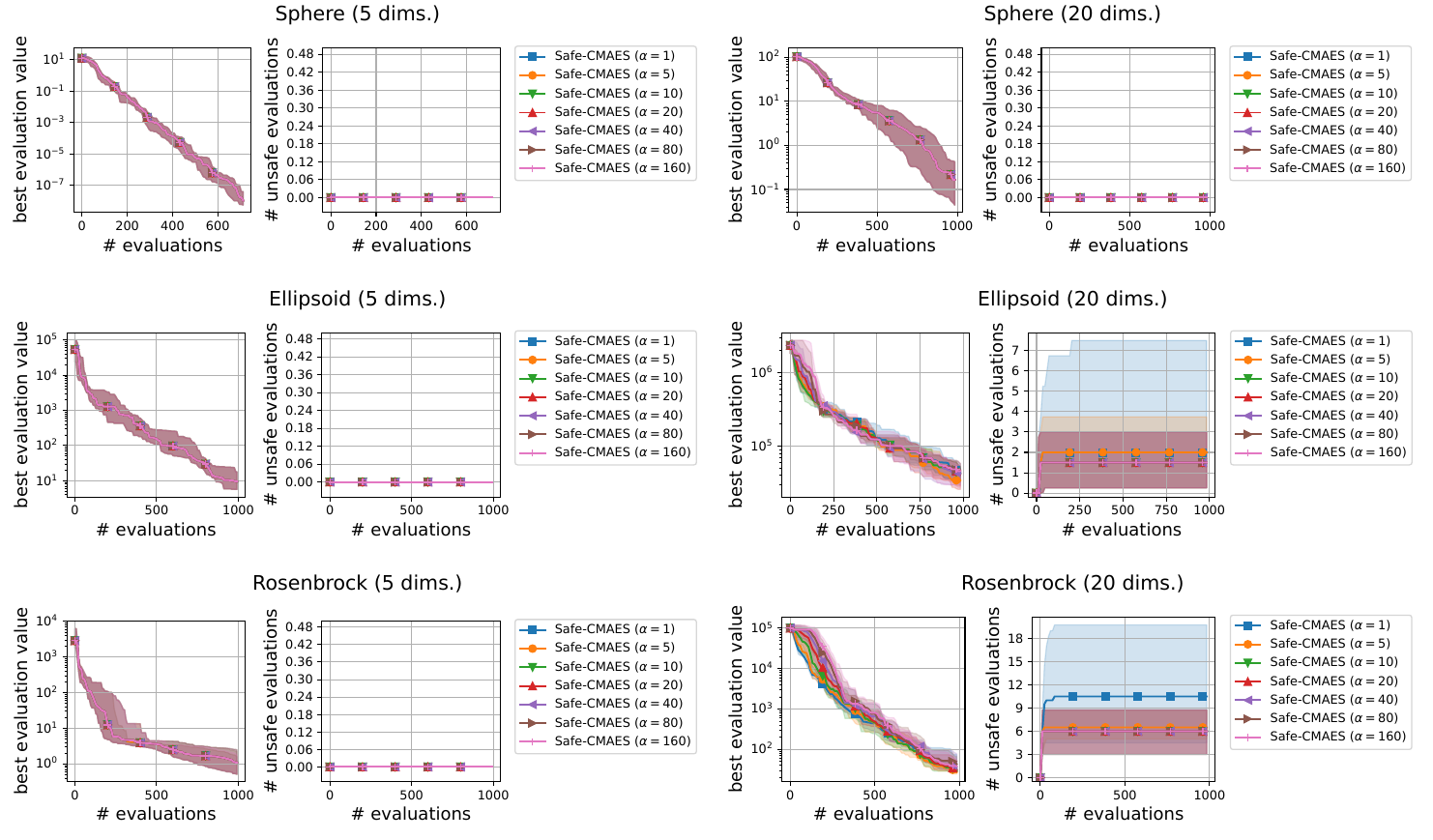}
    \caption{Result of Sensitivity Experiment of $\alpha$. We plot the medians and interquartile ranges of the best evaluation value and the number of evaluations of unsafe solutions.}
    \label{fig:alpha}
\end{figure*}
\begin{figure*}[t]
    \centering
    \includegraphics[width=0.9\linewidth]{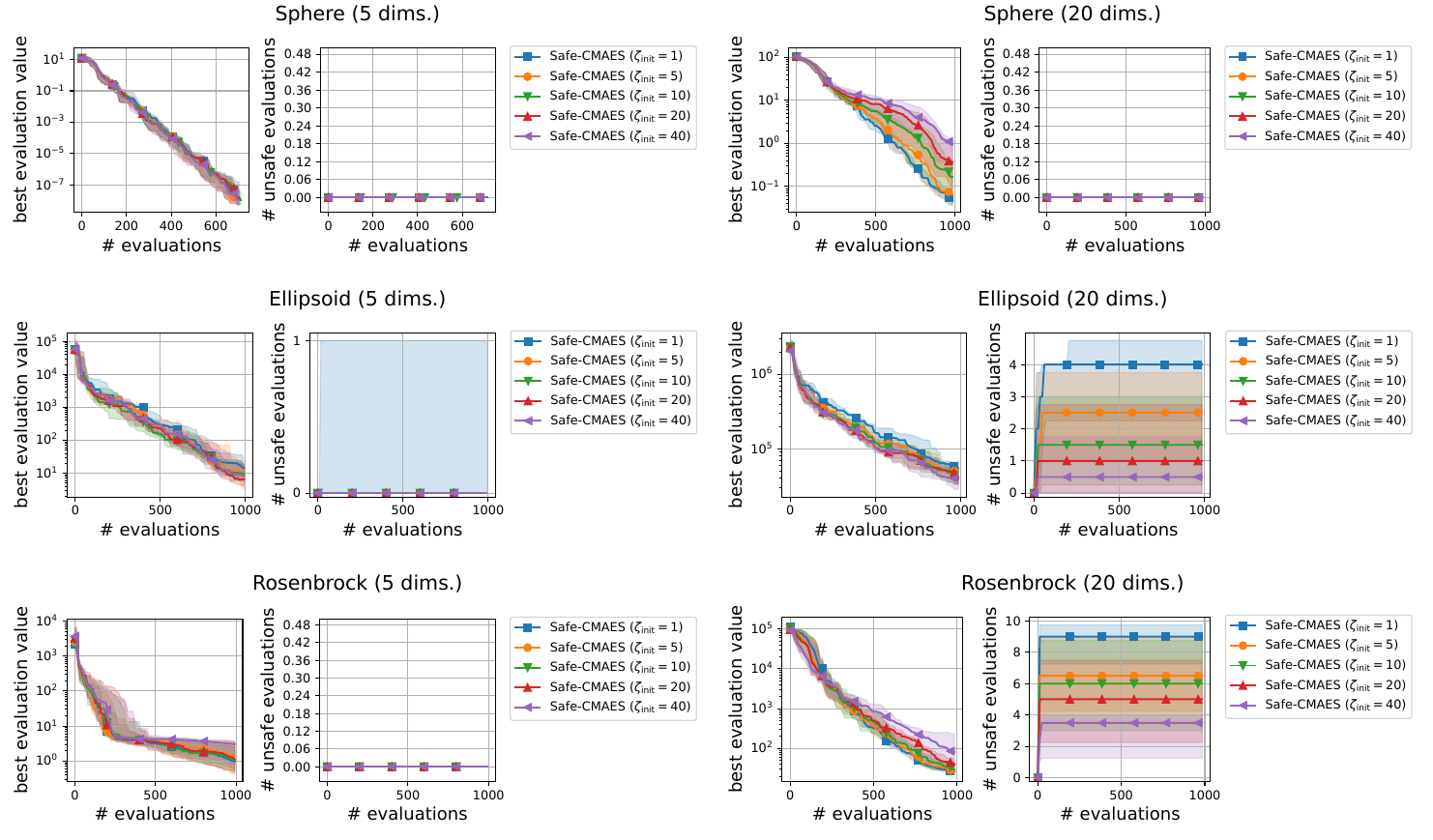}
    \caption{Result of Sensitivity Experiment of $\zeta_\mathrm{init}$. We plot the medians and interquartile ranges of the best evaluation value and the number of evaluations of unsafe solutions.}
    \label{fig:zeta}
\end{figure*}
\begin{figure*}[t]
    \centering
    \includegraphics[width=0.9\linewidth]{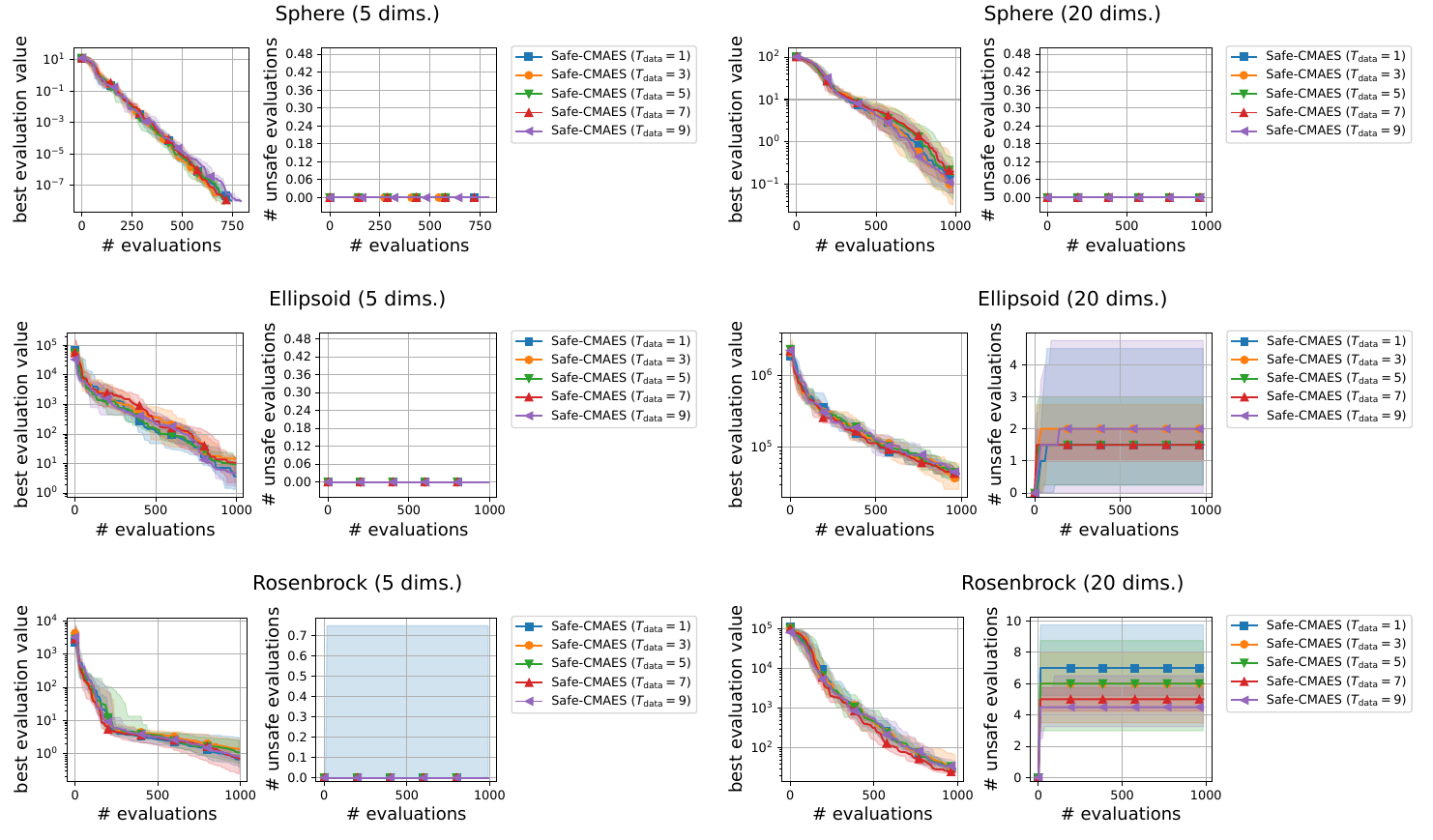}
    \caption{Result of Sensitivity Experiment of $T_\mathrm{data}$. We plot the medians and interquartile ranges of the best evaluation value and the number of evaluations of unsafe solutions.}
    \label{fig:Tdata}
\end{figure*}
%
%
%

\section{Investigation of Hyperparameter Sensitivity}
We investigated the sensitivities of hyperparameters $\alpha$, $\zeta_\mathrm{init}$, and $T_\mathrm{data}$ of the safe CMA-ES. 
The effect of those hyperparameters are summarized as
\begin{itemize}
    \item The hyperparameter $\alpha$ controls the increase and decrease rate of the coefficient $\rho_j^{(t+1)}$ for the estimated Lipschitz constant when the unsafe solutions are evaluated.
    \item The hyperparameter $\zeta_\mathrm{init}$ controls the coefficient $\tau^{(t+1)}$ for the estimated Lipschitz constant when the number of evaluated solutions used for the Gaussian process regression is small.
    \item The hyperparameter $T_\mathrm{data}$ controls the number of training data for Gaussian process regression.
\end{itemize}

We used the safety constraint used in the first and second experiments in the paper as
\begin{align}
    s(\x) = f(\x) \quad \text{and} \quad h = q(f, \X, 0.5) \enspace.
\end{align}
We run the safe CMA-ES changing a single hyperparameter and remaining the other hyperparameters to their recommended settings.
We ran 10 independent trials for each setting.

\subsection{Result of Sensitivity Experiment of $\alpha$}
We varied the setting of $\alpha$ as $\alpha = 1, 5, 10, 20, 40, 80, 160$.
Figure~\ref{fig:alpha} shows the transitions of the best evaluation value and the number of unsafe evaluations.
We note that the unsafe evaluation was not occurred in 5-dimensional functions and 20-dimensional sphere function.
Because the hyperparameter $\alpha$ affects the dynamics only when the unsafe evaluation occurred, the dynamics of the safe CMA-ES was not changed in those cases.
On the 20-dimensional ellipsoid function, the unsafe evaluations tended to be reduced as the hyperparameter $\alpha$ is large, while the best evaluation value was stagnated in the first few updates.
On the 20-dimensional rosenbrock function, the stagnation of best evaluation value was also confirmed with large $\alpha$, while the number of unsafe evaluations was not changed except for the case $\alpha = 1$, i.e., the case without the adaptation of coefficient $\rho_j^{(t+1)}$.
Totally, the recommended setting $\alpha = 10$ seems to be reasonable setting.

\subsection{Result of Sensitivity Experiment of $\zeta_\mathrm{init}$}
We varied the setting of $\zeta_\mathrm{init}$ as $\zeta_\mathrm{init} = 1, 5, 10, 20, 40$.
Figure~\ref{fig:zeta} shows the transitions of the best evaluation value and the number of unsafe evaluations.
When $\zeta_\mathrm{init} = 1$, i.e., when the adaptation of coefficient $\tau^{(t+1)}$ is not applied, the unsafe evaluation was occurred on 5-dimensional problems.
On the 20-dimensional sphere function, larger setting of $\zeta_\mathrm{init}$ delayed the decease of the best evaluation value while unsafe evaluation was not occurred in all setting.
In contrast, on the 20-dimensional ellipsoid function, the larger setting of $\zeta_\mathrm{init}$ reduced the unsafe evaluations maintaining the decrease rate of the best evaluation value.
We consider the performance of safe CMA-ES to be a bit sensitive to $\zeta_\mathrm{init}$, and the suitable setting of $\zeta_\mathrm{init}$ depends on the problem.

\subsection{Result of Sensitivity Experiment of $T_\mathrm{data}$}
We varied the setting of $T_\mathrm{data}$ as $T_\mathrm{data} = 1, 3, 5, 7, 9$.
Figure~\ref{fig:Tdata} shows the transitions of the best evaluation value and the number of unsafe evaluations.
The setting of $T_\mathrm{data}$ did not change the dynamics of the safe CMA-ES significantly on most of the problems.
When $T_\mathrm{data} = 1$, the unsafe evaluation was occurred on the 5-dimensional ellipsoid and rosenbrock functions.
On the 20-dimensional rosenbrock function, the setting $T_\mathrm{data} = 7$ most reduced the unsafe evaluations.
The recommended setting $T_\mathrm{data} = 5$ also works well in all problems.